\documentclass{article}

\usepackage{arxiv}

\usepackage[utf8]{inputenc} % allow utf-8 input
\usepackage[T1]{fontenc}    % use 8-bit T1 fonts
\usepackage{hyperref}       % hyperlinks
\usepackage{url}            % simple URL typesetting
\usepackage{booktabs}       % professional-quality tables
\usepackage{amsfonts}       % blackboard math symbols
\usepackage{nicefrac}       % compact symbols for 1/2, etc.
\usepackage{microtype}      % microtypography
\usepackage{lipsum}		% Can be removed after putting your text content
\usepackage{graphicx}
\usepackage{natbib}
\usepackage{doi}
\usepackage{makecell}
\usepackage{orcidlink}
\usepackage[titletoc]{appendix}

\graphicspath{ {./images/} }

\title{Iterative Multi-Agent Reinforcement Learning:\\A Novel Approach Toward Real-World Multi-Echelon Inventory Optimization}

\date{} 					% Or removing it

\author{ {\orcidlink{0009-0002-4807-5725}\hspace{1mm}Georg Ziegner} 
	\And
	{\orcidlink{0009-0002-6177-3974}\hspace{1mm}Michael Choi}
	\And
	{\orcidlink{0009-0001-1746-7315}\hspace{1mm}Hung Mac Chan Le}
	\And
	{\orcidlink{0009-0006-5687-8439}\hspace{1mm}Sahil Sakhuja}
	\And
	{\orcidlink{0009-0002-9212-5494}\hspace{1mm}Arash Sarmadi}
}

\date{}

\renewcommand{\headeright}{}
\renewcommand{\undertitle}{A Capstone Report in the Field of Data Science\\
for the Degree of Master of Liberal Arts in Extension Studies\\
Harvard University}
\hypersetup{
pdftitle={Iterative Multi-Agent Reinforcement Learning},
pdfsubject={A Novel Approach Toward Real-World Multi-Echelon Inventory Optimization},
pdfauthor={Georg Ziegner, Michael Choi, Hung Mac Chan Le, Sahil Sakhuja, Arash Sarmadi},
pdfkeywords={Data Science, Supply Chain Management, Reinforcement Learning},
}

\begin{document}
\maketitle

\begin{abstract}
	Multi-echelon inventory optimization (MEIO) is critical for effective supply chain management, but its inherent complexity can pose significant challenges. Heuristics are commonly used to address this complexity, yet they often face limitations in scope and scalability. Recent research has found deep reinforcement learning (DRL) to be a promising alternative to traditional heuristics, offering greater versatility by utilizing dynamic decision-making capabilities. However, since DRL is known to struggle with the curse of dimensionality, its relevance to complex real-life supply chain scenarios is still to be determined. This paper investigates DRL’s applicability to MEIO problems of increasing complexity. A state-of-the-art DRL model was replicated, enhanced, and tested across 13 supply chain scenarios, combining diverse network structures and parameters. To address DRL’s challenges with dimensionality, additional models leveraging graph neural networks (GNNs) and multi-agent reinforcement learning (MARL) were developed, culminating in the novel iterative multi-agent reinforcement learning (IMARL) approach. IMARL demonstrated superior scalability, effectiveness, and reliability in optimizing inventory policies, consistently outperforming benchmarks. These findings confirm the potential of DRL, particularly IMARL, to address real-world supply chain challenges and call for additional research to further expand its applicability. 
\end{abstract}

% keywords can be removed
% \keywords{First keyword \and Second keyword \and More}

\section{Introduction}
\paragraph{}
Warehouses play a critical role in supply chain management by serving as buffers against fluctuations in demand and supply. Storing inventory between the time it is procured and when it is distributed enables businesses to mitigate the risk of stockouts and better meet customer demand. Although businesses aim to avoid stockouts and lost sales, they also seek to prevent excess inventory, which increases carrying costs, ties up working capital, and may lead to stock obsolescence. By determining how much stock to maintain and how much to replenish at any moment, inventory policies can help strike the right balance between holding excess inventory and risking stockouts. \cite{arnold2008} % (Arnold et al., 2008)

\paragraph{}
While most businesses use inventory policies today, these are frequently used in a siloed manner within individual warehouses, leading to inventory levels that are optimized locally. Since many supply chains include multiple stages (echelons) through which inventory must flow before reaching the customer, MEIO seeks to manage inventory holistically throughout the entire supply chain, resulting in more efficient inventory management policies. \cite{geevers2024} % (Geevers et al., 2024)

\paragraph{}
In contrast to single-echelon inventory optimization (SEIO), a field that has been thoroughly researched and allows for mathematical models with closed-form solutions, MEIO models can be quite intricate, with solutions that are hard to determine. A typical approach to managing this complexity involves using heuristics, which can produce near-optimal solutions quickly and efficiently. However, the drawback is that heuristics are generally limited to narrowly defined supply chain situations, making their application to real-world scenarios more challenging \cite{gijsbrechts2022}. % {(Gijsbrechts et al., 2022)}

\paragraph{}
Reinforcement learning (RL) is a potentially more versatile approach to coping with the complexity of real-world MEIO problems. Designed to solve sequential decision-making problems in dynamic environments, RL can be used to find inventory policies for various supply chain scenarios. With the help of neural networks, RL can be enhanced into DRL, a powerful method for solving highly complex decision-making problems \cite{sutton2018}. % (Sutton \& Barto, 2018).

\paragraph{}
Recent research has shown that DRL can solve MEIO problems and that the resulting inventory policies can be more efficient than those derived from commonly used heuristics. However, the MEIO scenarios in which DRL has been tested often have limited complexity \cite{geevers2024}. % (Geevers et al., 2024).
Therefore, it remains to be determined whether DRL can be successfully applied to supply chain problems of real-life complexity.

\paragraph{}
In this paper, we applied DRL to increasingly complex multi-echelon inventory scenarios to test its applicability to more realistic supply chain problems. We began by replicating a state-of-the-art DRL model for MEIO based on existing research, which we modified for enhanced performance and flexibility. This modified model served as our base model, and we tested its performance across 13 different supply chain scenarios. To assess the base model’s performance, we compared it to a widely used heuristic, which we generalized to make it applicable to all scenarios in scope.

\paragraph{}
Building on our base model, we developed three additional models utilizing techniques from GNNs and MARL. After comparing the performance of these models with that of our base model, our conclusions led us to propose IMARL as a novel technique for solving MEIO problems of high complexity.

\paragraph{}
The remainder of this paper is organized as follows: In Chapter 2, we review the literature on solving MEIO problems using DRL. Chapter 3 details the methodology we employed for conducting experiments on various supply chain scenarios with five DRL models. In Chapter 4, we present the results of our experiments. Finally, Chapter 5 discusses the conclusions and offers suggestions for future research.

\section{Literature Review}
\paragraph{}
This chapter examines the current literature on inventory optimization and the methods available to solve this problem. We begin by discussing the concept of multi-echelon inventory management and the challenges stemming from its inherent complexity. Next, we explore common heuristics as a traditional approach to addressing this complexity. We then detail reinforcement learning and its advantages over heuristics before considering its applicability to the multi-echelon inventory management problem and its scalability constraints. Finally, we introduce several enhancements to traditional reinforcement learning to mitigate these scalability issues.

\subsection{Multi-Echelon Inventory Management}
\paragraph{}
With inventory usually representing 20\% to 60\% of a manufacturing company’s total assets (Arnold et al., 2008), effective control of the invested working capital through inventory management offers an important potential for improvement. Since Harris developed the economic order quantity (EOQ) in 1913, many researchers have taken an interest in modeling and analyzing inventory systems \cite{gumus2007}. % (Gümüs \& Güneri, 2007).

\begin{figure}
  \centering
  \includegraphics[width=\textwidth,height=\textheight,keepaspectratio]{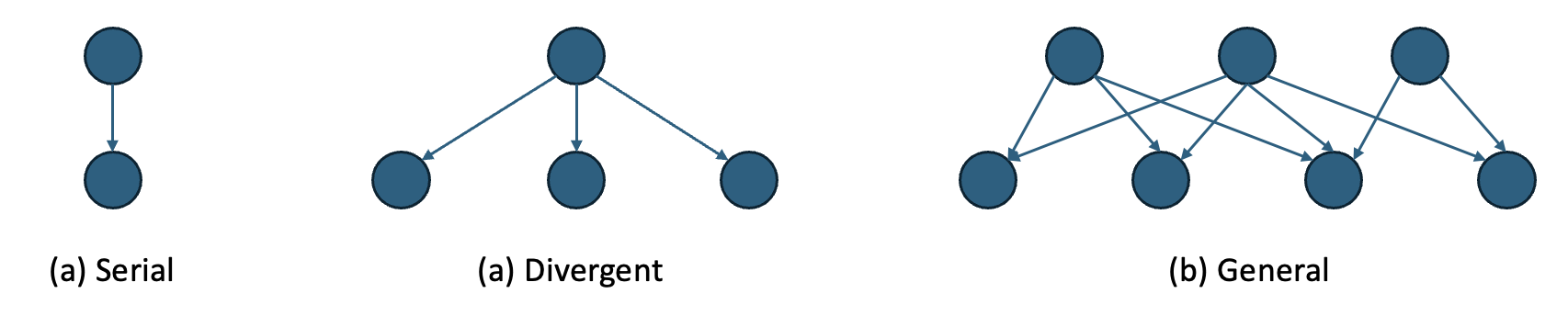}
  \caption{Three Types of Inventory Network:\\\textit{Example structures for the primary types of inventory networks: serial, divergent, and general.}}
  \label{fig:fig1}
\end{figure}

\paragraph{}
The term ‘multi-echelon’ refers to supply chains where stock locations—such as production facilities, warehouses, distribution centers, and retailers—are organized in a network of multiple stages through which products must pass before reaching the final customer. The structure of these networks can vary in complexity. In serial inventory networks, inventory flows linearly, with each stock point supplied by one upstream location and supplying one downstream location. In divergent inventory networks, at least one stock point feeds into multiple downstream locations. Finally, in general inventory networks, at least one stock point is supplied by multiple upstream locations, and at least one stock point supplies multiple downstream locations \cite{geevers2024}. %(Geevers et al., 2024).
Figure \ref{fig:fig1} provides a graphical representation of the three inventory network types.

\paragraph{}
The multiple interdependent stages that characterize multi-echelon inventory systems make them significantly more challenging to manage than single-echelon systems. This is one reason why many businesses continue to manage the inventory of individual stock points independently, considering only immediate upstream supply and downstream demand. However, since this approach overlooks the interdependencies between adjacent echelons, it can result in certain stock points holding excess inventory while others often face shortages \cite{li2013}. %(Li, 2013).
According to \cite{vandeput2020}, %Vandeput (2020),
applying MEIO can reduce inventory by 10\% to 35\% compared to SEIO.

\paragraph{}
Various research techniques have been employed to tackle the complexity of managing multi-echelon inventory systems, with a focus on mathematical models and heuristics \cite{gumus2007}. 
%(Gümüs & Güneri, 2007).
Although exact mathematical models have been successfully developed for simpler supply chain structures, they often are too complex and time-consuming to solve more intricate supply chain problems due to the curses of dimensionality \cite{geevers2024}.%(Geevers et al., 2024).

\subsection{Heuristics for Inventory Management}
\paragraph{}
Heuristics are a practical alternative for managing multi-echelon inventory systems when mathematical models are intractable. Although these heuristics can quickly find near-optimal solutions, their applicability is usually limited to specific inventory systems and network parameters \cite{gijsbrechts2022}. % (Gijsbrechts et al., 2022).

\paragraph{}
The most commonly used heuristics in inventory management are base-stock policies \cite{geevers2024}. %(Geevers et al., 2024). 
They ensure a constant supply to meet demand by placing sufficient replenishment order quantities to maintain a constant base-stock level \cite{clark1960} %S. Clark and Scarf (1960) 
proved that base-stock policies are optimal for serial inventory systems, and their decomposition method allows efficient computation of the optimal base-stock level S. Based on this work, \cite{shang2003} %Shang and Song (2003) 
proposed heuristics that provide closed-form solutions to the base-stock levels in serial inventory networks. Subsequently, \cite{rong2017} %Rong et al. (2017) 
have further enhanced this approach by introducing their decomposition-aggregation (DA) heuristic, which can be used to derive near-optimal base-stock levels for divergent networks. This paper extends the DA heuristic to general inventory networks, stochastic shipment lead times, and complex demand distributions. We used this extended heuristic as a benchmark for our experiments.

\subsection{(Deep) Reinforcement Learning}
\paragraph{}
RL is a method to solve complex sequential decision-making problems. In RL, agents interact with an environment to maximize a cumulative reward. In each time step, agents observe the current state of the environment, based on which they choose an action and receive a reward, while the environment transforms into a new state. This problem is known as a Markov decision process (MDP), and RL can be used to solve it and find optimal policies \cite{sutton2018}.%(Sutton & Barto, 2018).

\setlength{\parindent}{0pt}

\paragraph{}
Early RL algorithms represent the environment using tabular methods or shallow function approximators. While this is practical for environments with small discrete action and state spaces, these algorithms struggle when the state or action space is large or high-dimensional. To scale RL to more complex environments, \cite{mnih2015} %Mnih et al. (2015) 
proposed DRL, a method that uses deep neural networks to approximate complex values or policy functions.

\paragraph{}
Value-based RL methods, such as Deep Q-Network (\cite{mnih2015}), %(Mnih et al., 2015), 
derive a policy by approximating the value of the state-action pairs. In contrast, policy-based RL methods, such as actor-critic methods (\cite{sutton2018}), %(Sutton & Barto, 2018), 
directly approximate the policies. The latter achieve better results for high-dimensional or continuous action spaces but often with a less stable performance. Hybrid algorithms like proximal policy optimization (PPO) (\cite{schulman2017}) %(Schulman et al., 2017) 
combine the benefits of both value-based and policy-based algorithms to achieve better results.

\subsection{Reinforcement Learning for MEIO}
\paragraph{}
Inventory management can be modeled as an RL problem with the environment representing the supply chain. The RL agent interacts with the environment by taking actions determining the quantity and timing of replenishment orders. Depending on the business goal, the reward signal can be derived from the supply chain’s cost, profit, or service level.

\paragraph{}
Furthering the work by \cite{oroojlooyjadid2022} %Oroojlooyjadid et al. (2022) 
that demonstrated the value of using DRL in inventory optimization, \cite{geevers2024} %Geevers et al. (2024) 
have conducted recent research into this topic with an extensive review of existing approaches for using DRL for MEIO. They deduced that most approaches rely on extensive assumptions and simplifications that are quite different from real-world scenarios. To address this, they applied DRL to multi-echelon inventory systems using an approach designed to resemble real-life supply chains more closely. Firstly, they applied a continuous action space, thereby avoiding the limitations associated with discrete action spaces used in most research. Secondly, their model was designed to be versatile and could be applied to different inventory network structures. Finally, they are among the first researchers to use DRL for inventory management on general network structures, which are more representative of actual supply chains.

\paragraph{}
While the approach developed by \cite{geevers2024} %Geevers et al. (2024) 
can be considered state-of-the-art, it also relies on simplifications that limit its applicability for solving real-world MEIO problems. For their experiments, they assumed static shipment lead times and simple demand distributions. Although the modeled inventory networks are more complex than those used in most other research, they include only two echelons and a very limited number of stock points. Moreover, their model struggled with scalability issues, as it did not converge reliably when dealing with complex inventory networks.

\subsection{Enhancements to RL}
\paragraph{}
When DRL algorithms are applied to more complex environments with large state and action spaces, the curse of dimensionality can cause scalability issues, impacting the algorithm’s effectiveness. Several techniques can be used to reduce the dimensionality of the state and action spaces.

\paragraph{}
Compared to single-agent reinforcement learning (SARL), where only one agent interacts with the environment, MARL decomposes large decision problems into smaller ones that are solved by multiple agents. While MARL is known to reduce model complexity and improve convergence speed in high-dimensional spaces, it creates several new challenges. Non-stationarity is introduced as agents continuously change their policies during their learning process, causing other agents to adapt to these changes, which can lead to cyclic and unstable learning behavior. Another challenge lies in multiple equilibriums that can be reached when agents update their policies in parallel. In this case, agents need to agree on one of these equilibriums, which is non-trivial and can result in a suboptimal global policy. Finally, allocating rewards to individual agents can be challenging, as it is not always clear how specific agents and their actions contributed to an outcome (\cite{albrecht2024}). %(Albrecht et al., 2024). Yu et al. (2022) 
\cite{yu2022} showed promising results from multi-agent proximal policy optimization (MAPPO) for agents in a cooperative setting where the critic network is shared among agents. %Papoudakis et al. (2021)
\cite{papoudakis2021} showed that MAPPO, in general, outperforms independent proximal policy optimization (IPPO), where each agent has its own actor and critic networks. In this project, we utilized MAPPO to implement our MARL-based models.

\paragraph{}
Based on the work by %Bertsekas (2021), Wang et al. (2023) 
\cite{bertsekas2021}, \cite{wang2023} proposed a technique to mitigate the non-stationarity challenge of MARL by updating agents’ policies sequentially rather than simultaneously. Their method essentially turns non-stationary MARL problems into stationary SARL problems, resulting in a strong performance when tested on popular MARL problems. The IMARL model proposed in this paper uses a similar approach.

\paragraph{}
GNNs are specifically designed to work with graph structures and effectively reduce model dimensionality by encoding node and edge features to lower-dimensional embeddings. However, as GNNs are computationally expensive to run, dimensionality reduction with GNNs often creates new scalability challenges (\cite{liu2020}).%(Liu & Zhou, 2020).

\paragraph{}
In MARL scenarios, GNNs can also capture relational information among agents and help overcome information-sharing constraints. When combining GNNs with MARL to solve inventory management problems, \cite{kotecha2024} %Kotecha and Chanona (2024) 
achieved promising results by leveraging the inherent graph structure of a supply chain to support collaboration between its entities.

\section{Methodology}
\paragraph{}
As discussed in Chapter 2, applications of DRL to MEIO used in current research often rely on extensive assumptions and simplifications, limiting their applicability to real-life supply chain problems. We ran a series of experiments to explore whether DRL can solve MEIO problems of real-life complexity. In this chapter, we describe the RL environment that we used to simulate the supply chain. Next, we present our research scope, consisting of a set of supply chain scenarios of increasing complexity before introducing the algorithm and experiment configuration. After that, we describe five DRL models used to run our experiments and the heuristic that serves as our performance benchmark. We close with remarks on data analysis techniques, tools used to run the experiments, and ethical considerations.

\subsection{RL Environment}
\paragraph{}
Our starting point for modeling the MEIO problem was the model developed by \cite{geevers2024}, %Geevers et al. (2024), 
which we modified in terms of its state function and sequence of events. We closely follow their approach and terminology with some adjustments to their notation when describing our RL environment.

The environment contains a set of stock points $p \in P$. To indicate a direct upstream or downstream relationship between a stock point $p$ and other entities in the network, we denote its upstream predecessors (including external suppliers) $u \in U_{p} \in P$ and its downstream successors (including external customers) $d \in D_{p} \in P$. While all stock points share the same notation, we may refer to stock points with internal customers as warehouses and those with external customers as retailers. The full notation is provided in Table \ref{tab:table1}. The problem is formulated as an MDP with an RL agent acting as a central decision maker. At each period $t$, the agent observes the state $s_t$ (\ref{eq:1}) and takes the action $a_t$. The state contains the inventory position (IP) for each stock point at period $t$.

\begin{equation}
  \label{eq:1}
  s_t = IP_{p,t} \quad \forall p \in P
\end{equation}

\begin{table}
  \caption{Notation}
   \centering
   \resizebox{\textwidth}{!}{%
      \begin{tabular}{ll}
     \toprule
     %\multicolumn{2}{l}{Indices} \\
     $Indices$ & \\
     \cmidrule(r){1-1}
     $p$      & Stock points (warehouses and retailers), $p \in P$ \\
     $u$      & Upstream predecessors (internal and external suppliers) of stock point $p$, $u \in U_{p} \in P$ \\
     $d$      & Downstream successors (internal and external customers) of stock point $p$, $d \in D_{p} \in P$ \\
     $t$      & Time period, $t = 1,2, \dots T$ \\
     %\midrule
     & \\
     $Parameters$ & \\
     \cmidrule(r){1-1}
     $h_p \geq 0$ & Holding cost per period for stock point $p$ \\
     $b_p \geq 0$ & Backorder cost per period for stock point $p$ \\
     $\lambda_{u,p,t} \geq 0$ & Shipment lead time from upstream stock point $u$ to stock point $p$ in period $t$\\
     $q_{p,t} \geq 0$ & External demand at stock point $p$ in period $t$ \\
     & \\
     $Variables$ & \\
     \cmidrule(r){1-1}
     $I^{begin}_{p,t} \in \mathbb{Z}$ & Inventory level at stock point $p$ in period $t$ after receiving incoming shipments \\
     $I^{end}_{p,t} \in \mathbb{Z}$ & \makecell[l]{Inventory level at stock point $p$ in period $t$ after sending outgoing shipments \\ to downstream entities and external customers} \\
     $BO_{p,d,t} \in \mathbb{Z}$ & Backorders of stock point $p$ to downstream location $d$ in period $t$ \\
     $IT_{p,u,t} \in \mathbb{Z}$ & Number of items in transit from upstream entity $u$ to stock point $p$ to arrive in period $t$ \\
     $O_{p,u,t} \in \mathbb{Z}$ & Number of items ordered by stock point $p$ from upstream entity $u$ in period $t$ \\
     $IP_{p,t} \in \mathbb{Z}$ & Inventory position of stock point $p$ in period $t$ \\
     $BSL_{p} \in \mathbb{Z}$ & Base-stock level of stock point $p$ \\
     \bottomrule
    \end{tabular}%
   }
   \label{tab:table1}
   \\
   \raggedright\textit{A list of indices, parameters, and variables used throughout the study to define the RL environment for MEIO.}
  \end{table}

\paragraph{}
IP (\ref{eq:2}) is an important summary statistic in inventory management. It refers to the amount of inventory available or expected to be available, factoring in on-hand inventory, inventory on order, and inventory in transit. While \cite{geevers2024} %Geevers et al. (2024) 
used the individual IP components in their state variable, we found that reducing the state to a summary statistic yields better results.

\begin{equation}
  \label{eq:2}
  IP_{p,t} = I_{p,t} + \sum_{u \in U_p}\left[ O_{p,u,t} + BO_{u,p,t} \right] - \sum_{d \in D_p}\left[ O_{d,p,t} + BO_{p,d,t} \right]
\end{equation}

\paragraph{}
The action (\ref{eq:3}) taken at a given time determines stock points’ replenishment order quantities. While stock points located in the most upstream echelon place replenishment orders to external suppliers with unlimited supply, all other stock points order from their internal upstream suppliers in the network. If a stock point has multiple upstream suppliers, it places its entire order amount for a given period to only one of them based on random choice.

\begin{equation}
  \label{eq:3}
  a_t = O_{p,u,t} \quad \forall p \in P
\end{equation}

\begin{table}
\renewcommand{\arraystretch}{1.5}
  \caption{Sequence of Events Describing the MDP Transition Function }
   \centering
   \resizebox{\textwidth}{!}{%
      \begin{tabular}{ l l }
     \toprule

     \textbf{Event 1} \qquad Receive incoming shipments \\
     $I^{begin}_{p,t} = I^{end}_{p,t-1} + IT_{u,p,t}$ \\
     \midrule
     \textbf{Event 2}    \qquad Fulfill downstream orders, prioritizing backorders before handling new orders \\
     $I'_{p,t} = I^{begin}_{p,t}$\\
     $BO_{p,d,t} = IT_{p,d,t+ \lambda_{p,d,t}} = 0$ \\
     \textbf{for} \ $Order_{p,d,t-1}$ \textrm{in} \ $[BO_{p,d,t-1}, \ O_{d,p,t-1}]:$ \\
     \qquad \textbf{for} \ $d$ \ \textrm{in ascending order of} \ $IP_{d,t}:$ \\
     \qquad\qquad$\textbf{if} \ Order < I'_{p,t} \ \textbf{then}:$ \\
     \qquad\qquad\qquad$IT_{p,d,t + \lambda_{p,d,t}} = Order_{p,d,t-1}$ \\
     \qquad\qquad\qquad$BO_{p,d,t} = 0$ \\
     \qquad\qquad\qquad$I'_{p,t} = I'_{p,t} - Order_{p,d,t-1}$ \\
     \qquad\qquad$\textbf{else}:$ \\
     \qquad\qquad\qquad$IT_{p,d,t + \lambda_{p,d,t}} = I'_{p,t}$ \\
     \qquad\qquad\qquad$BO_{p,d,t} = Order_{p,d,t-1} - I'_{p,t}$ \\
     \qquad\qquad\qquad$I'_{p,t} = 0$ \\
     $I^{end}_{p,t} = I'_{p,t}$ \\
     % \multicolumn{2}{l}{$I'_{p,t} = I^{begin}_{p,t}$} \\
     % \multicolumn{2}{l}{$\textbf{for} \ Order_{p,d,t-1} in [BO_{p,d,t-1}, O_{d,p,t-1}]:$} \\
     % \multicolumn{2}{l}{\qquad$\textbf{for} d in ascending order of IP_{d,t}:$} \\
     % \multicolumn{2}{l}{\qquad\qquad$\textbf{if} Order < I'_{p,t} \textbf{then}:$} \\

     % \multicolumn{2}{l}{\qquad\qquad\qquad$IT_{p,d,t + \lambda_{p,d,t}} = Order_{p,d,t-1}$} \\
     % \multicolumn{2}{l}{\qquad\qquad\qquad$BO_{p,d,t} = 0$} \\
     % \multicolumn{2}{l}{\qquad\qquad\qquad$I'_{p,t} = I'_{p,t} - Order_{p,d,t-1}$} \\
     % \multicolumn{2}{l}{\qquad\qquad$\textbf{else}$} \\
     % \multicolumn{2}{l}{\qquad\qquad\qquad$IT_{p,d,t + \lambda_{p,d,t}} = I'_{p,t}$} \\
     % \multicolumn{2}{l}{\qquad\qquad\qquad$BO_{p,d,t} = Order_{p,d,t-1} - I'_{p,t}$} \\
     % \multicolumn{2}{l}{\qquad\qquad\qquad$I'_{p,t} = 0$} \\
     % \multicolumn{2}{l}{$I^{end}_{p,t} = I'_{p,t}$} \\
     %\cmidrule{1-2}
     \midrule
     \textbf{Event 3} \qquad Upstream orders are placed based on action (for warehouse stock points with internal customers) \\ and external demand (for retailer stock points with external customers) \\
     $a{t}, q_{p,t} \to O_{p,u,t}$ \\
     \bottomrule
    \end{tabular}%
   }
   \label{tab:table2}
   \\
   \raggedright\textit{A detailed sequence outlining the transition events in the MDP used to model inventory dynamics. It includes receiving shipments, fulfilling orders, and placing new upstream orders.}
 \end{table}

\paragraph{}
Once the action is taken, the state transitions into a new state $s_{t+1}$ with a probability determined by the transition function $\Delta(s_{t+1} | a_t, s_t)$. The sequence of events (Table \ref{tab:table2}) determines the transition function with randomness added through uncertainty of external demand $q_{p,t}$ and shipment lead times $\lambda_{u,p,t}$ between the upstream entity and the receiving downstream stock point.

\paragraph{}
Every time step $t$ is characterized by three sequential events. First, stock points receive shipments scheduled to arrive in the current period $t$ from upstream locations ($IT_{u,p,t}$). Next, stock points attempt to fulfill downstream orders placed by internal and external customers in the previous period ($O_{p,d,t-1}$) using their on-hand inventory. Where possible, they send outgoing shipments ($IT_{p,d,t+\lambda_{p,d,t}}$) with shipment lead time $\lambda_{p,d,t}$. If there is not enough on-hand inventory to fulfill all orders, downstream stock points are ranked based on their IP ($IP_{d,t}$), and those with the lowest IP are served first until no inventory is left. In case stockouts occur, the unfulfilled order amount is backordered ($BO_{p,d,t}$) and will be fulfilled with priority over new incoming orders in the following period. Finally, incoming orders ($O_{d,p,t}$) are placed at all stock points, determined either by the action $a_t$ (for warehouses) or by randomly generated external demand $q_{p,t}$ (for retailers). It should be noted that our second event, the fulfillment of downstream orders, is carried out in parallel for all network echelons rather than sequentially, as defined by \cite{geevers2024}. %Geevers et al. (2024). 
We chose to deviate from their approach, as we consider parallel fulfillment to be more closely aligned with supply chain events in real life.

\paragraph{}
Transitioning from state $s_t$ to $s_{t+1}$ as a result of action $a_t$ generates a reward $R(s_t,a_t,s_{t+1})$. Since our goal is to minimize the total cost of the supply chain, which includes holding costs $h_p$ and backorder costs $b_p$ for the stock points $p \in P$, the reward is defined by the negative costs, as described by the cost function (\ref{eq:4}).

\begin{equation}
  \label{eq:4}
  c_t(s_t,a_t) = \sum_{p \in P}\left[h_p \cdot I_{p,t} + b_p \cdot \sum_{u \in U_p} BO_{p,u,t} \right]
\end{equation}

\paragraph{}
A policy $\pi$ can be defined by mapping states to actions. Our objective is to find the optimal policy $\pi^{\ast}$ (\ref{eq:5}) (\cite{geevers2024})
for which the sum of costs, discounted by factor $\gamma$, is minimized over a large period $T$.

\begin{equation}
  \label{eq:5}
  \pi^{\ast} = arg \min_{\pi \in \Pi} \mathbb{E} \left[ \sum^{\infty}_{t = 0} \gamma^t \cdot c_t(s_t,a^{\pi}_t) \right]
\end{equation}

\subsection{Research Scope}

\paragraph{}
We designed a complexity grid to test DRL’s applicability to MEIO problems of varying complexity. Here, we considered network structure and network parameters as two distinct drivers of supply chain complexity relevant to our experiments. We defined levels of complexity for both drivers and combined them into a set of complexity scenarios.

\subsubsection{Network Structure}
\begin{figure}
  \centering
  \includegraphics[width=\textwidth,height=\textheight,keepaspectratio]{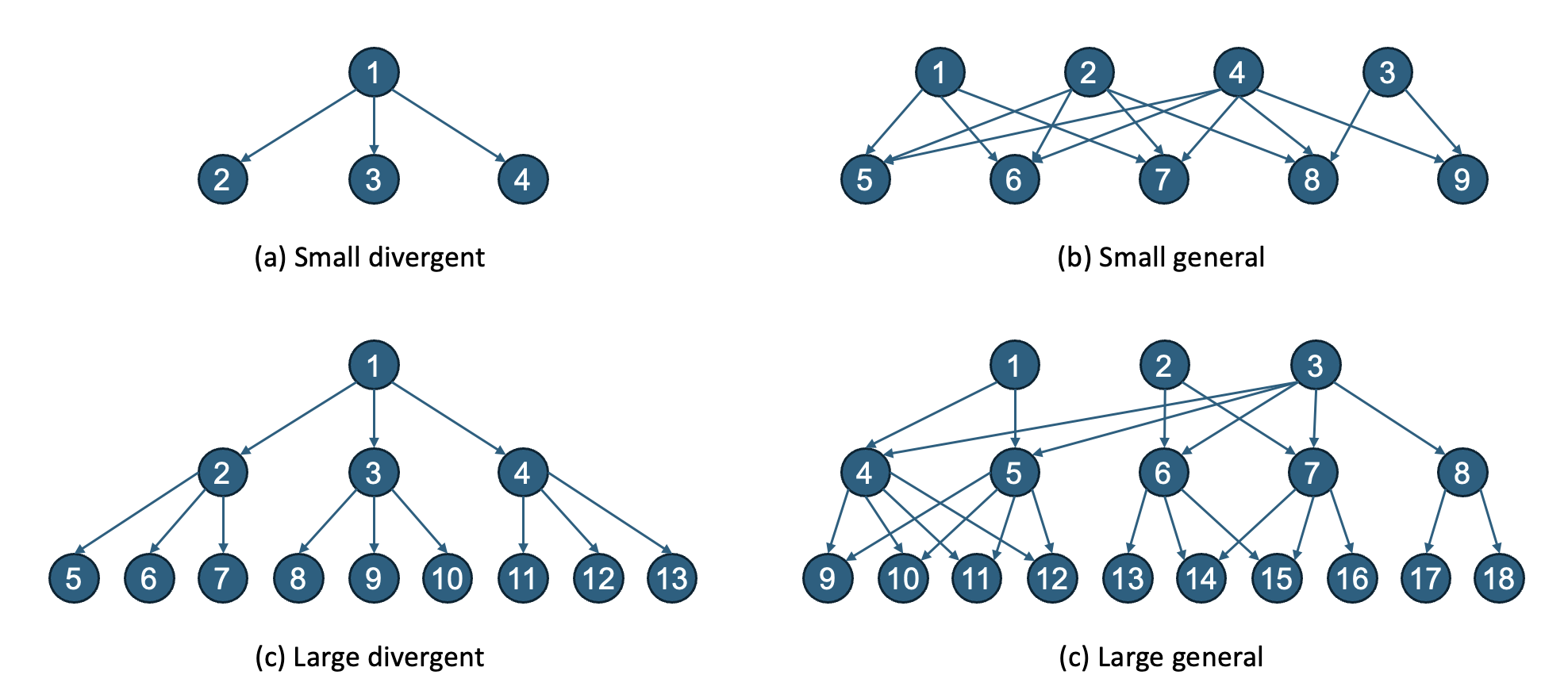}
  \caption{Modeled Network Structures\\\textit{Visual depiction of the four network structures used in the experiments, differentiated by the number of stock points and echelons, as well as the type of inventory network (divergent or general). }}
  \label{fig:fig2}
\end{figure}

\paragraph{}
Network structure defines the upstream-to-downstream relationship between all stock points within the supply chain. Its complexity is mainly driven by its type (i.e., serial, divergent, and general) and depth (i.e., the number of echelons). As serial supply chains are more straightforward and have been subject to extensive research, we decided to exclude them from our research scope and instead focus on the following four structures depicted in Figure \ref{fig:fig2}.

\begin{enumerate}
\item Small divergent: 4 stock points arranged in 2 echelons 
\item Small general: 9 stock points arranged in 2 echelons 
\item Large divergent: 13 stock points arranged in 3 echelons 
\item Large general: 18 stock points arranged in 3 echelons 
\end{enumerate}

\subsubsection{Network Parameters}
\begin{figure}
  \centering
  \includegraphics[width=\textwidth,height=\textheight,keepaspectratio]{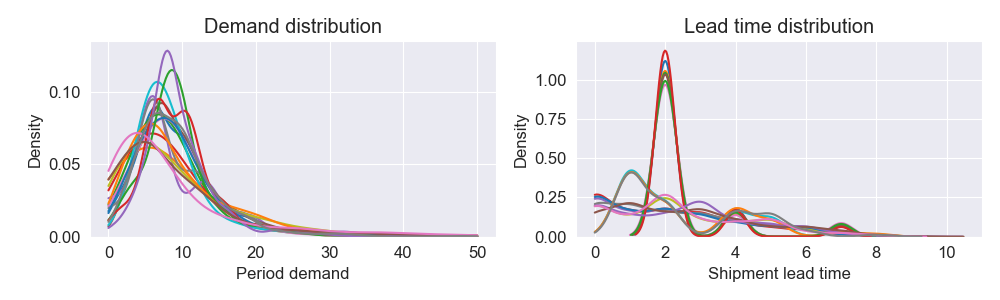}
  \caption{Real-Life Demand and Shipment Lead Time Density Plots\\\textit{Probability density plots visualizing the variability in real-life demand and shipment lead times. The plots represent the first 18 products in the dataset (Yang et al., 2023b), corresponding to the maximum number of stock points modeled in the network structures.}}
  \label{fig:fig3}
\end{figure}

\paragraph{}
Our RL environment contains three types of parameters: external demand, shipment lead time, and costs. The complexity of the MEIO problem increases with lower predictability of these parameters for a given simulation period. While in the existing research, network parameters are often assumed to be static or to follow simple probability distributions, they tend to be highly stochastic in real-life supply chains. We ran our experiments for combinations of the following complexity levels:

\paragraph{External Demand}
Following the approach used by \cite{geevers2024}, %Geevers et al. (2024), 
we started by generating individual demand for each retailer and each period from a Poisson distribution with a mean drawn from a Uniform distribution ($q_{p,t} \sim \textrm{Pois(Unif\{$5,15$\})}$). To better represent real-life demand, we also utilized a dataset (\cite{yang2023a}, \cite{yang2023b}) %(Yang et al., 2023a, 2023b) 
representing daily retail demand for 2778 products over 40 months. By converting the data for each of these products into a probability mass function (after scaling it to a mean of 10), we created a set of random variables representing daily demand. To simulate different demand patterns across the network, each stock point in our simulated supply chain was allocated a different random variable to generate demand. Density plots of these demand distributions are available in Figure \ref{fig:fig3}.

\paragraph{Shipment Lead Time}
We applied three levels of complexity when modeling lead time. For the simplest scenario, we assumed a static lead time of $\lambda_{u,p,t} = 1$ between all stock points and their upstream suppliers. Lead time becomes stochastic in the following scenario, where we sample it from a Uniform distribution for all edges in the network graph ($\lambda_{u,p,t} \sim \textrm{Unif\{$1,5$\}}$). While these first two scenarios replicate the approach used by \cite{geevers2024}, %Geevers et al. (2024), 
our third scenario is based on actual lead time information from the same dataset that we used to simulate real-life demand (\cite{yang2023a}, \cite{yang2023b}). %(Yang et al., 2023a, 2023b). 
Using the same approach, we again generated random variables from data for 2778 retail products (after scaling it to a mean of 3) and simulated shipment lead times for all shipments made in a period. Density plots for the lead times are shown in Figure \ref{fig:fig3}.

\paragraph{Costs}
Our model contains two types of costs: holding cost $h_p$ and backorder cost $b_p$. While we had to use static cost values to keep the scope of this project manageable, we acknowledge that introducing stochastic costs to the model forms an important next step in making the simulation even more realistic. Moreover, our simulations do not consider any fixed ordering costs (e.g., shipping fees), whose introduction to the model would have a significant impact on the resulting policies. We chose to exclude them from the scope of this project as they would require a different type of heuristic as a performance benchmark, which does not scale as well to the different complexity scenarios as the heuristic described later in this chapter.

\subsubsection{Complexity Grid Framework}
\begin{figure}
  \centering
  \includegraphics[width=\textwidth,height=\textheight,keepaspectratio]{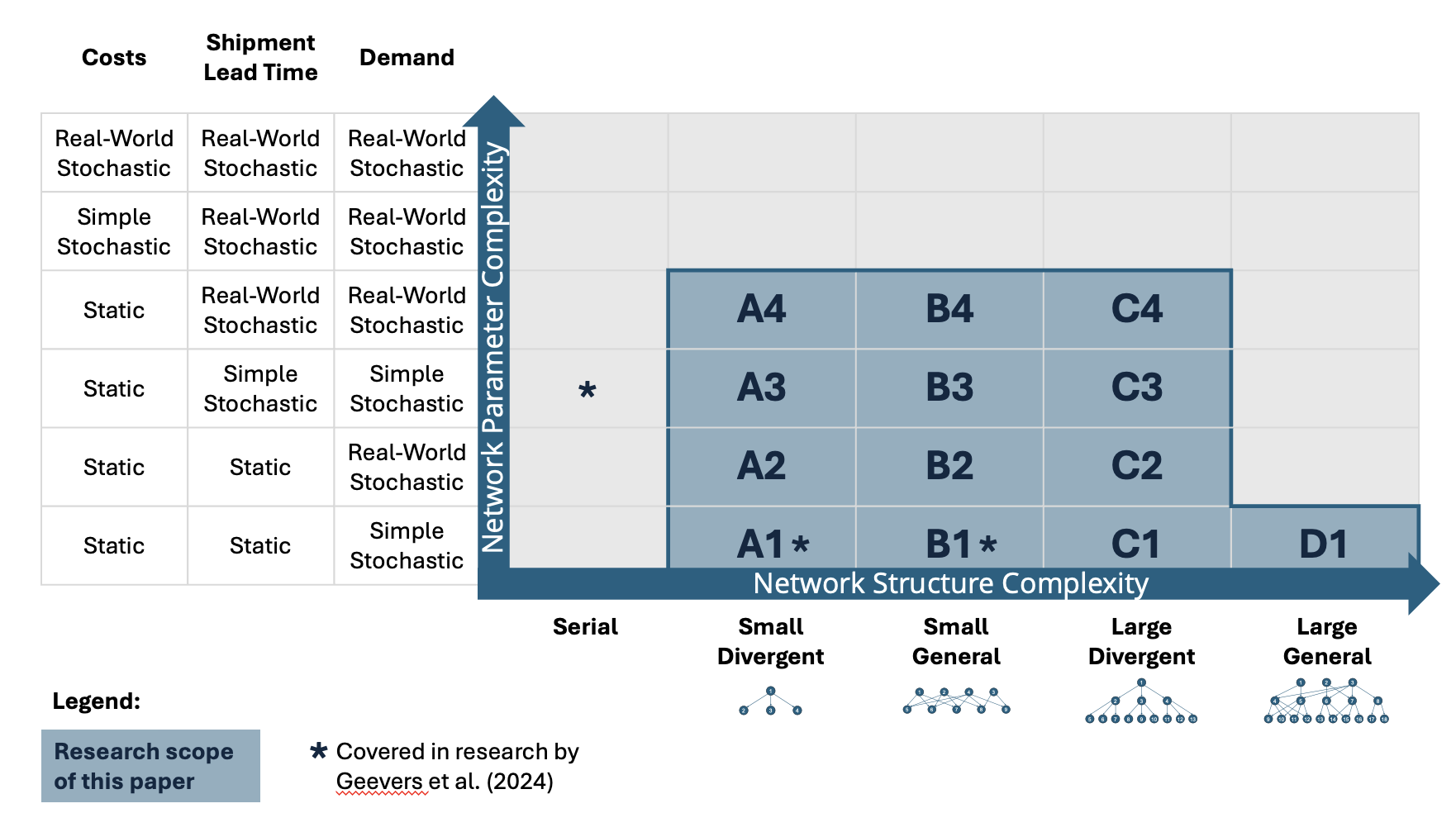}
  \caption{Complexity Grid Framework\\\textit{A structured framework mapping the complexity of supply chain scenarios by network structure and parameters, highlighting the scenarios included in this study and those covered in previous research by Geevers et al. (2024).}}
  \label{fig:fig4}
\end{figure}

\paragraph{}
We combined the previously described complexity levels for the network structures and parameters in a complexity grid framework (Figure \ref{fig:fig4}), mapping 30 distinct scenarios, of which we selected 13 for running our experiments. Apart from serial network structures and stochastic network costs, we also excluded three scenarios relating to the large general network structure. This decision was made considering the limited computing resources available to us at the time of writing.

\subsection{Algorithm and experiment configuration}
\paragraph{}
In the previous sections, we introduced 13 supply chain scenarios representing our research scope and the RL environment used for the simulation. This section will present the algorithm and configuration used to train our models.

\subsubsection{Algorithm}
\paragraph{}
Following the approach by \cite{geevers2024}, %Geevers et al. (2024), 
we used PPO (\cite{schulman2017}) %(Schulman et al., 2017) 
to train our models to solve the MDP. While \cite{geevers2024} %Geevers et al. 
used the standard implementation of PPO, we have found slightly better results for a different set of hyperparameters, which we provide in Appendix B. Also, adapting our PPO implementation from the PureJaxRL library (\cite{lu2023}) %(Lu et al., 2023) 
allowed us to run multiple vectorized environments in parallel.

\subsubsection{Parameters and Configuration}
\paragraph{}
All experiments across the 13 complexity scenarios followed a common setup in which we ran episodes consisting of 128 time steps in the MDP. At the beginning of each episode, we initialized the environment by setting inventory levels for all stock points to the base-stock levels derived from the benchmark heuristic while setting orders, backorders, and in-transit inventory to zero. This approach helped to avoid unfavorable episode starts, where initial orders cannot be fulfilled due to missing inventory, resulting in high backorder quantities early on.

\paragraph{}
With increasing scenario complexity, the solution space grows, and more training iterations are needed to find good policies. Therefore, we adapted the total number of training episodes based on the supply chain scenario and the DRL model. At every 100 training episodes, we simulated the policy learned at that point by performing evaluation runs of 100 episodes consisting of 75 time steps each. For each of these evaluation runs, we collected and reported the cumulative reward over episodes’ last 50 time steps, considering the initial 25 time steps a warm-up period.

\paragraph{}
As indicated in the previous section, most network parameters change with the complexity scenario to allow different levels of predictability (Table \ref{tab:table3}). However, for holding cost and backorder cost, we used static parameters. Backorder costs are incurred only by retail stock points facing external demand. Holding costs exist for all stock points and strictly increase from upstream to downstream echelons. These conditions were used in the heuristic by \cite{shang2003} %Shang and Song (2003) 
and thus formed a requirement for our performance benchmark to be valid. Table \ref{tab:table4} summarizes the cost parameters applied in our simulations.

\paragraph{}
To address model scalability, we used continuous state and action spaces, which grow linearly with the number of stock points in our simulated supply chain. With continuous state and action variables, appropriate boundaries are required that limit the problem and solution spaces without making them too small to support exploration. These boundaries are echelon-specific, and we list the values applied for each complexity scenario in Table \ref{tab:table4}.

\paragraph{}
To foster algorithm convergence and reduce its training time, we applied several transition functions on the input and output of our RL environment. Actions were scaled up from $[-1, 1]$ to $[0, O^{max}_{p,u,t}]$ and state observations were scaled down from $[IP^{min}_{p,t}, IP^{max}_{p,t}]$ to $[-1, 1]$. Also, based on \cite{geevers2024}, %Geevers et al. (2024), 
we divided the reward by 1000 before passing it to the actor-critic network.

\begin{table}
  \caption{Complexity Scenario-Specific Parameters}
   \centering
   \resizebox{0.6\textwidth}{!}{%
    \begin{tabular}{ c c c c }
     \toprule
     Scenario     &     Network Structure       &       $q_{p,t}$         &       $\lambda_{u,p,t}$ \\
     \cmidrule{1-4}
     A1     &     Small Divergent       &       Pois(Unif\{$5,15$\})         &       $1$ \\
     A2     &     Small Divergent       &       Real-life data         &       $1$ \\
     A3     &     Small Divergent       &       Pois(Unif\{$5,15$\})         &       Unif\{$1,5$\} \\
     A4     &     Small Divergent       &       Real-life data         &       Real-life data \\
     B1     &     Small General       &       Pois(Unif\{$5,15$\})         &       $1$ \\
     B2     &     Small General       &       Real-life data         &       $1$ \\
     B3     &     Small General       &       Pois(Unif\{$5,15$\})         &       Unif\{$1,5$\} \\
     B4     &     Small General       &       Real-life data         &       Real-life data \\
     C1     &     Large Divergent       &       Pois(Unif\{$5,15$\})         &       $1$ \\
     C2     &     Large Divergent       &       Real-life data         &       $1$ \\
     C3     &     Large Divergent       &       Pois(Unif\{$5,15$\})         &       Unif\{$1,5$\} \\
     C4     &     Large Divergent       &       Real-life data         &       Real-life data \\
     D1     &     Large General       &       Pois(Unif\{$5,15$\})         &       $1$ \\
     \bottomrule
    \end{tabular}%
   }
   \label{tab:table3}
  \\
  \raggedright\textit{Parameter values defining demand and shipment lead time distributions for each complexity scenario, highlighting the progression from static to real-world stochastic conditions.}
\end{table}%

\begin{table}
  \caption{Echelon-Specific Parameters and Bounds}
    \centering
    \resizebox{\textwidth}{!}{%
    \begin{tabular}{ l c c c c c c c }
      \toprule
      Echelon     & $h_p$ & $b_p$ & $O^{min}_{p,u,t}$ & $O^{max}_{p,u,t}$  & $IP^{min}_{p,t}$ & $IP^{max}_{p,t}$ & Applicable Complexity Scenarios\\
      \cmidrule{1-8}
      1 (retailer) & $1$ & $19$ & $0$ & $50$  & $-100$ & $BSL_p + 50$ & all \\
      2 & $0.6$ & $0$ & $0$ & $150$  & $-300$ & $BSL_p + 150$ & all \\
      3 & $0.4$ & $0$ & $0$ & $500$  & $-1000$ & $BSL_p + 500$ & C1, C2, C3, C4, D1 \\
      \bottomrule
    \end{tabular}%
    }
    \label{tab:table4}
  \\
  \raggedright\textit{Holding and backorder cost parameters, along with upper and lower bounds for state and action variables, specified for each echelon in the modeled supply chains. Echelons are numbered from downstream to upstream, with echelon 1 representing the retailer nodes that face external demand.}
\end{table}

\subsection{Tested Models}
\paragraph{}
Next, we will introduce the five models we trained to solve the 13 supply chain problems in our research scope.

\subsubsection{Model 1: SARL (Base Model)}
\paragraph{}
Our first model uses SARL, where one agent acts as the central decision-maker within the RL environment. Accounting for the deviations in the experimental setting and algorithm configuration, this model is similar to the one used by \cite{geevers2024} %Geevers et al. (2024) 
to successfully solve MEIO problems as indicated in Figure \ref{tab:table4}, and is thus considered our base model.

\subsubsection{Model 2: SARL+GNN}
\paragraph{}
The next model adds a GNN to the base model, with supply chain stock points as nodes and relationships between stock points as directed edges in the graph. The node features are given by that part of the environment state that is directly related to that node. We applied multiple Graph Attention Network (GAT) layers that use attention scores to consider the features of a node’s neighbors when transforming that node’s own features into an embedding. This embedding was then used as input to the critic network.

\subsubsection{Model 3: MARL}
\paragraph{}
In our MARL model, $n$ agents representing $n$ stock points in the network solve the MDP in a decentralized way by choosing actions that directly relate to their respective stock points. Likewise, agents only observe that part of the environment’s state representing their network node. While for retail locations, agents’ rewards depend only on their respective locations, other agents receive the total network reward, as their actions impact the network beyond their own stock point. Based on these design choices, retail stock points are homogeneous in terms of state, action, and reward, as well as their parameters (i.e., demand, shipment lead time, and costs). For this reason, we trained agents representing retail locations using a single actor network, resulting in a shared policy. All other agents were trained individually with their own actor network, as their policies depend on their respective locations in the network. The critic network is shared among all agents to foster collaboration.

\subsubsection{Model 4: MARL+GNN}
\paragraph{}
In analogy to the previous two models, this model uses the same GNN-based embedding as in model 2 as input to the central critic network designed for model 3. 

\subsubsection{Model 5: IMARL}
\paragraph{}
Adding to the previous four models, which have been researched in the context of MEIO to various degrees (\cite{geevers2024}; \cite{kotecha2024}' \cite{yang2023a}), %(Geevers et al., 2024; Kotecha & Chanona, 2024; Yang et al., 2023a), 
we propose IMARL as a novel DRL model. By applying a multi-agent approach and thereby reducing problem dimensionality, IMARL addresses the scalability issues of SARL while mitigating the non-stationarity challenges of MARL through an iterative training approach.

\paragraph{}
Similar to model 3, every stock point of the supply chain network in IMARL is represented by an agent that takes actions for that stock point in a decentralized way. Deviating from the MARL model, agents observe an environment state that not only represents their own stock point but also those located downstream and thus are impacted by them. Similarly, agents’ rewards are based on the costs incurred by their respective stock points and all the entities located downstream from them.

\paragraph{}
Unlike in regular MARL, where all agents are trained in parallel, IMARL agents are trained one at a time for a certain number of episodes, iterating over agents until a stable solution is found. At the start, all agents are initialized with a base policy before being trained for the first time. We used the benchmark heuristic as base policy to accelerate the learning process. While one agent is being trained, all other agents deterministically apply their policy, allowing the one agent to act in a stationary and low-dimensional problem space. As IMARL iterates over agents, their base policies are successively replaced by the policies they learned during their training. In further training iterations after that, agents update their policies only if they learn a better policy during that training iteration. Whenever an agent updates its policy with a better one, all its direct upstream predecessors and downstream successors in the network also get a chance to update their policies. This process continues until no agents can find better policies, meaning that an equilibrium is found. Table \ref{tab:table5} describes IMARL’s training logic using pseudocode.

\begin{table}
  \caption{IMARL Training Logic as Pseudocode}
    \centering
    \resizebox{0.75\textwidth}{!}{%
    \begin{tabular}{ l }
      \toprule
      \texttt{randomly initialize saved\_policy for all agents} \\
      \texttt{initialize job\_list containing all agents} \\
      \textbf{while} \texttt{job\_list is} \textbf{not} \texttt{empty:} \\
      \qquad \texttt{remove first agent from job\_list and define it as active\_agent} \\
      \qquad \textbf{for} \texttt{n episodes:} \\
      \qquad \qquad \textbf{for} \texttt{agent in untrained non-active agents:} \\
      \qquad \qquad \qquad \texttt{act deterministically using base policy} \\
      \qquad \qquad \textbf{for} \texttt{agent in trained non-active agents:} \\
      \qquad \qquad \qquad \texttt{act deterministically using agent's saved\_policy} \\
      \qquad \qquad \texttt{train active agent using PPO based on saved\_policy} \\
      \qquad \texttt{define best\_policy across n training episodes} \\
      \qquad \textbf{if} \texttt{best\_policy is better than active\_agent's saved\_policy} \textbf{then}: \\
      \qquad \qquad \texttt{replace saved\_policy with best\_policy} \\
      \qquad \qquad \texttt{add active\_agent's predecessors and successors to job\_list} \\ 
      \bottomrule
    \end{tabular}%
    }
    \label{tab:table5}
  \\
  \raggedright\textit{Pseudocode describing IMARL’s high-level training logic. Iteration resumes for as long as agents find superior policies, which results in their direct predecessors and successors being retrained.}

\end{table}

\subsection{Benchmark Heuristics} 
\paragraph{}
To assess the quality of the inventory policies derived from our models, we compared their performance against that of a benchmark heuristic. As the base-stock policy is known to perform well under the conditions in our simulated supply chain, our task was to find a heuristic that can derive near-optimal base-stock levels for all stock points. Since inventory management heuristics usually depend on a specific supply chain setup, we struggled with finding a heuristic that is valid for all the complexity scenarios in our research scope. The DA heuristic (\cite{rong2017}), %(Rong et al. 2017), 
which was used by \cite{geevers2024} %Geevers et al. (2024) 
to benchmark the performance of their models, is defined for divergent network structures facing Poisson demands and static shipment lead times. When exploring the DA heuristic, we found that it offers the potential to be extended to other demand types, stochastic lead times, and general network structures.

\paragraph{}
The DA heuristic consists of three steps: First, the divergent network is decomposed into a set of serial systems. Next, near-optimal base-stock levels are calculated for each stock point in the serial system using the heuristic from \cite{shang2003}. %Shang and Song (2003). 
Finally, the serial systems are aggregated back into the original network by applying a “backorder matching” procedure. In this procedure, a stock point’s base-stock level is set to match the sum of expected backorders for its counterparts in all serial systems.

\paragraph{}
The last two steps in the heuristic rely on the lead time demand distribution, which describes the variation in product demand during the time it takes to replenish inventory and is given by the convolution of the demand distribution and the lead time distribution. In a supply chain network, an upstream stock point’s total lead time demand distribution is the convolution of the individual lead time demand distributions from all its direct downstream stock points, assuming that lead time and demand are independent across these downstream stock points. In the DA heuristic, these convolutions can be derived with minimal effort due to the favorable properties of the Poisson distribution and the static lead times defined by \cite{rong2017}.% Rong et al. (2017).

\paragraph{}
To extend the DA heuristic to other demand distributions and stochastic lead times, we considered these distributions’ individual probability mass functions and manually performed convolutions on them. As this process quickly becomes computationally expensive due to the large number of permutations, we applied fast Fourier transforms to accelerate the process. Extending the DA heuristic to general distribution networks was possible, as in our simulation environment, stock points place their orders randomly to any of their upstream suppliers. Demand as a random variable is independent in this scenario, allowing us to calculate upstream stock points’ lead time demand distributions considering the probabilities of not receiving orders from some of their downstream locations in a given period.

\paragraph{}
By extending the DA heuristic to custom demand and lead time distributions, as well as to general network structures (assuming independent demand and lead time), this paper contributes to operations research. While we did not compare the performance of our extended heuristic to that of other more specialized heuristics and need to assume that specialized heuristics might perform better for some of our complexity scenarios, we believe that the extended DA heuristic serves as a highly sufficient benchmark for testing the performance of our DRL models.

\subsection{Tools Used}
\paragraph{}
Our initial implementation of the RL environment and models relied on Gymnasium, RLlib, and Stable Baselines3, which are widely used RL-specific Python libraries, and on stockpyl \cite{snyder2023}, % (Snyder, 2023), 
a library designed for inventory optimization and simulation. While these libraries support graphics processing unit (GPU) based accelerated processing, we found it to be up to 100\% slower than central processing unit (CPU) based training. This is a common challenge with RL, as the frequent interactions with the environment constitute an input/output bottleneck that is tighter than the computation of the model itself. Relying on CPU-based processing only, a single run of our base model took approximately 2.5 hours for the least complex scenario, A1, and multiple days for the most complex scenario, D1. Facing these computational times, it was unfeasible for us to test different improvement ideas, and we had to change our approach.

\paragraph{}
As a result, we changed our implementation to rely on JAX. This high-performance Python library provides the ability to compile and run parallel just-in-time optimized functions, which enables efficient GPU-based execution. As JAX relies on a functional programming approach built around pure functions, immutability, and composition, the transition to JAX required us to fully rebuild our implementation, including the supply chain process-specific elements previously handled by stockpyl. The transition to JAX resulted in significant speed improvements that reduced training run durations by 30 to 40 times, which enabled us to test different parameter combinations and explore more complex models systematically.

\subsection{Data Analysis Techniques}
\paragraph{}
In RL, agents learn by interacting dynamically with the environment and exploring new actions that, due to randomness, often lead to delayed and varying rewards. These factors contribute to a higher outcome variability in RL model training compared to supervised learning. To account for this variability, we performed ten separate runs, each initialized with a different random seed, for all our experiments.

\paragraph{}
As previously indicated, we assessed the quality of our models by comparing their performance with that of the benchmark heuristic. We were interested in a model’s potential effectiveness, as measured by its best performance across all runs, and its reliability, as measured by its average performance across runs.

\paragraph{}
Finally, we wanted to interpret the policies that our models have learned. To determine the actions a model would take for a stock point in a given state, we simulated the supply chain over a large number of time periods, recording both the state values and the corresponding actions taken by the trained model.

\subsubsection{Ethical Considerations}
\paragraph{}
In the context of inventory management, DRL is an emerging topic that has yet to be widely tested outside of academia. This work can be considered a proof of concept and, as such, is not intended for direct implementation in an actual business setting. For that reason, and since we are using either simulated or fully anonymized data, we did not adapt our methods based on ethical considerations. However, for future implementations intended to be used in actual supply chains, we would like to highlight the following considerations.

\subsubsection{Data Privacy, Security, and Integrity}
\paragraph{}
Inventory management systems often include sensitive data such as sales patterns, customer behavior, and supplier information. Safe storage and handling of this data are essential to address data privacy concerns and safeguard it against breaches. Special attention should also be paid to data integrity, as poor data quality can negatively affect the supply chain, customers, and partners.

\subsubsection{Transparency and Accountability}
\paragraph{}
In the context of inventory management, it is essential that supply chain operators understand the rationale behind decisions made by a DRL model, such as prioritizing specific stock points, suppliers, or products. If the model makes an error leading to overstocking or stockouts, operators should be able to challenge or override that decision. Clear accountability regarding who is responsible for that decision should be defined.

\subsubsection{Fairness and Social Impact}
\paragraph{}
Artificial intelligence tools often optimize decisions to maximize efficiency and minimize costs. As this may result in policies that can disadvantage small suppliers or impact product availability in developing regions, it is essential to design these tools so as not to disproportionally harm specific stakeholders in the supply chain. Similarly, it should be avoided to make the supply chain vulnerable to disruptions caused by dependencies on certain suppliers.

\subsubsection{Environmental Impact}
\paragraph{}
Each year, 8\% of global inventory is discarded due to expiry or overproduction, amounting to annual waste worth \$163 billion (\cite{avery2022}), %(Avery Dennison, 2022), 
more than the gross domestic products of Costa Rica, Croatia, and Iceland combined. By further optimizing inventory policies and reducing excess inventories and waste, DRL has the potential to make a significant positive contribution to the environment. However, prioritizing cost-effectiveness in inventory optimization can also lead to undesired environmental consequences, such as unnecessary emissions through higher shipment frequencies. Therefore, environmental sustainability should be assessed holistically and integrated into a model’s decision-making priorities.

\section{Analysis}
\paragraph{}
In the previous chapter, we introduced the methodology used to simulate supply chains with different levels of complexity and introduced five DRL models to solve the MEIO problem. We ran experiments to compare the performance of these methods by applying them to 13 different supply chain scenarios of increasing complexity. To account for the variability in results that is inherent to reinforcement learning, each experiment was repeated ten times.

\paragraph{}
In this chapter, we present the simulation results and compare the methods’ performance and consistency along the complexity grid framework. Next, we compare the models’ learning behavior and identify potential caveats that should be addressed in future research. Finally, we analyze the learned policies of the DRL methods and compare them to the heuristic that served as our benchmark.

\subsection{Experiment Results}
\paragraph{}
As explained in previous chapters, inventory optimization in multi-echelon supply chains today relies on heuristics that, while not necessarily optimal, produce good results quickly and efficiently. To assess the quality of a DRL method, we needed to compare its performance to that of inventory management heuristics. Only if DRL can consistently outperform heuristics in a specific supply chain scenario can it become relevant for businesses to solve MEIO problems in real life. For each trial, we consider the best reward a model achieved across all evaluation runs and define this as the trial’s result. For the ten trials we ran for each experiment, we report the best overall result, indicating the model’s effectiveness in solving the MEIO problem, and the average result, indicating the model’s reliability.

\subsubsection{Model Effectiveness}
\begin{figure}
  \centering
  \includegraphics[width=\textwidth,height=\textheight,keepaspectratio]{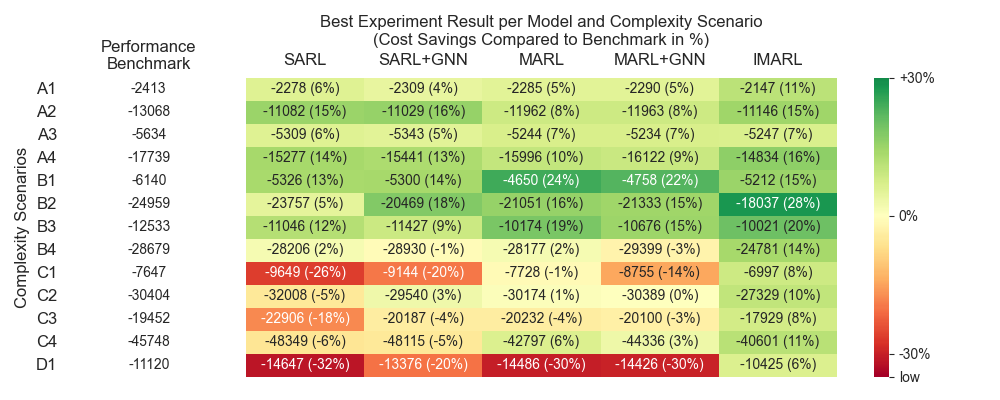}
  \caption{Best Experiment Results per Model and Complexity Scenario\\\textit{A heat map showing the best results achieved by each tested model across the 13 supply chain scenarios. Cell colors correspond to percentage values provided in parentheses after the absolute results, reflecting cost savings relative to the benchmark.}}
  \label{fig:fig5}
\end{figure}

\paragraph{}
For each experiment, we compared the models’ best result across ten trials with the benchmark heuristic result for the respective complexity scenario. The difference between these two indicates the cost savings that a model can generate compared to the benchmark heuristic in each scenario. These results are summarized in Figure \ref{fig:fig5}, which shows that all methods could beat the benchmark for simple supply chain scenarios. However, with increasing scenario complexity, fewer DRL methods outperformed the benchmark. The SARL baseline model adapted from \cite{geevers2024} %Geevers et al. (2024) 
started underperforming in supply chain scenarios with three echelons. While MARL outperformed SARL in most complexity scenarios, it still failed to surpass the benchmark in many of the more complex cases. The GNN-based models did not generally perform better or worse than their SARL and MARL counterparts, often showing minimal differences in results. A possible explanation lies in our supply chain design, where IP was the only node feature that GNNs could leverage to provide added value over other methods.
 
\paragraph{}
IMARL clearly stood out in the comparison, performing well in all simulated supply chain scenarios. Not only did it outperform the heuristic in all scenarios, but it also achieved the highest reward across the five methods in ten of these scenarios.

\subsubsection{Model Reliability}
\begin{figure}
  \centering
  \includegraphics[width=\textwidth,height=\textheight,keepaspectratio]{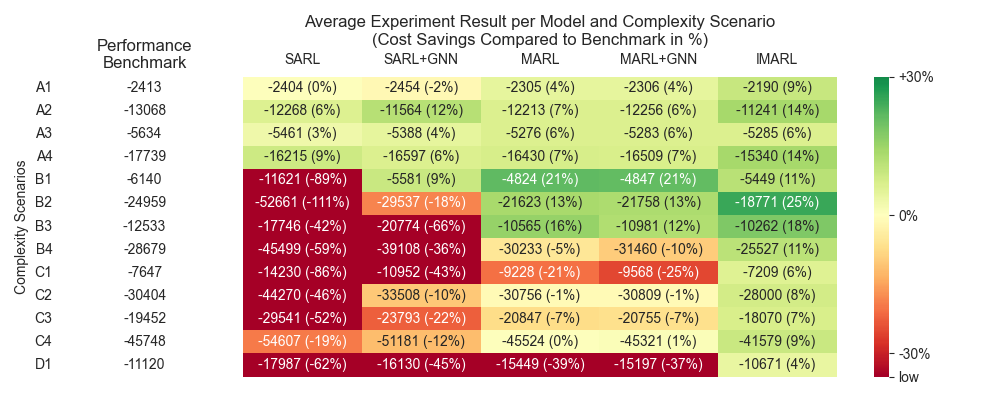}
  \caption{Average Experiment Results per Model and Complexity Scenario\\\textit{A heat map presenting the average performance of each model across ten trials for the 13 supply chain scenarios. Cell colors correspond to percentage values provided in parentheses after the absolute results, reflecting cost savings relative to the benchmark.}}
  \label{fig:fig6}
\end{figure}

\begin{figure}
  \centering
  \includegraphics[width=\textwidth,height=\textheight,keepaspectratio]{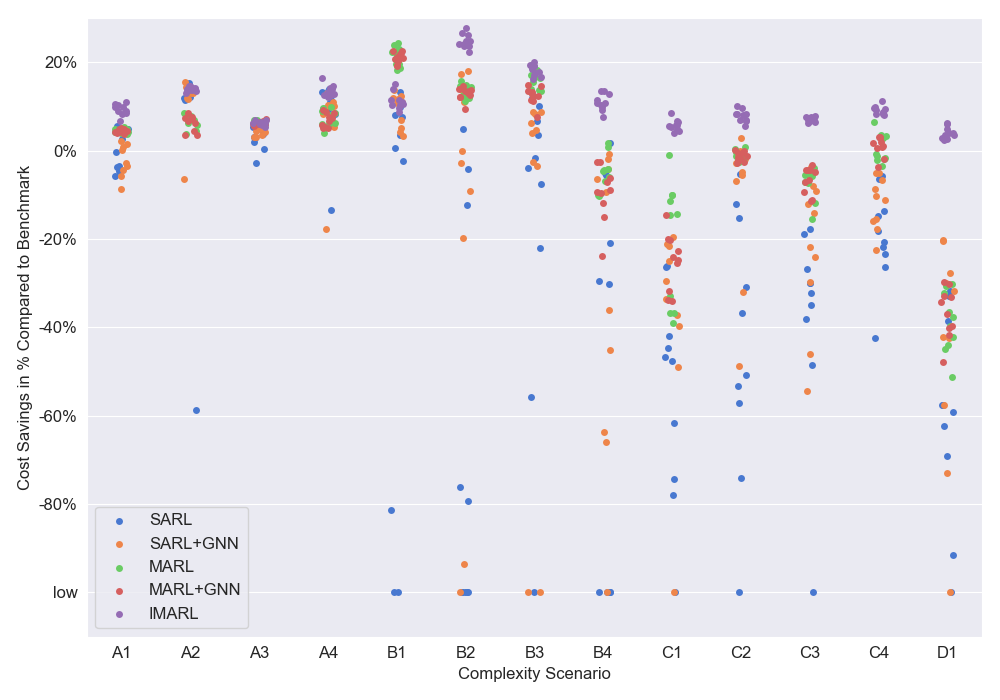}
  \caption{Experiment Trial Results per Model and Complexity Scenario\\\textit{Visualization of experiment trial results, illustrating the relative cost savings compared to the benchmark heuristic, as well as tested models' result variability.}}
  \label{fig:fig7}
\end{figure}

\paragraph{}
We assess a model’s reliability by its ability to produce similar results consistently. Figure \ref{fig:fig6} reports the average results across ten trials for each experiment, which indicates that many models that could beat the benchmark with their best result could not do so consistently. Figure \ref{fig:fig7} shows the individual trial results of all models relative to the benchmark, providing additional insights about the variability of model results.

\paragraph{}
Both models with single agents (SARL and SARL+GNN) exhibit greater variability than the three MARL-based models (MARL, MARL+GNN, and IMARL). While single-agent models could occasionally outperform the benchmark in low-complexity supply chain scenarios, they failed to do so consistently. When comparing the average experiment results, the single-agent models outperformed the benchmark in scenarios with small divergent network structures (A1–A4) but underperformed in most other scenarios. The multi-agent models performed consistently well in even more of the less complex supply chain scenarios (A1–B3), surpassing the benchmark in most trials. As scenario complexity increases, the MARL and MARL+GNN models also see higher variability in their results.

\paragraph{}
Again, IMARL set itself apart from other models by demonstrating remarkable consistency. Across 130 trials conducted in 13 scenarios, IMARL consistently outperformed the benchmark. As a result, its average performance exceeded the benchmark in every scenario, establishing IMARL as both an effective and reliable model.

\subsection{Learning Behavior and Interpretation of Results}
\paragraph{}
This section analyzes models’ learning curves to understand how each method trains on the data and how its learning behavior impacts its results. For IMARL, we also investigate potential limitations and dependencies.

\subsubsection{Learning Curves}
\paragraph{}
As indicated in the previous chapter, when running experiments on the five models, we evaluated model performance for 100 episodes after every 100 training episodes. The mean cumulative reward was recorded for each of these evaluation runs, and its development over the entire training duration forms the learning curve for an experiment. 

\begin{figure}
  \centering
  \includegraphics[width=\textwidth,height=\textheight,keepaspectratio]{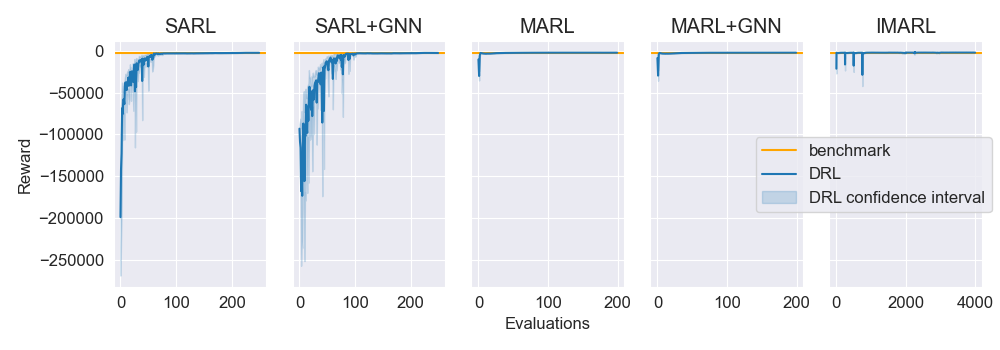}
  \caption{Example Learning Curves by Model\\\textit{Learning curves for models in complexity scenario B1, illustrating differences in training dynamics and convergence behavior.}}
  \label{fig:fig8}
\end{figure}

\paragraph{}
Figure \ref{fig:fig8} shows the learning curves of the five models for complexity scenario B1, which is representative of the models’ learning behavior across the complexity grid. Learning curves for other grid combinations are provided in Appendix C. Three distinct patterns could be observed when comparing models’ learning behavior.

\paragraph{}
Single-agent models (SARL and SARL+GNN) converged slowly but steadily toward a solution. Big jumps in model performance characterize the early stage of training, whereas the policies become more stable towards the end. This learning behavior results from the large problem and solution spaces that single agents must explore. As scenario complexity increases, so does the solution space, which makes it harder for a single agent to find good policies in a reasonable amount of time.

\paragraph{}
Multi-agent models with parallel training of agents (MARL and MARL+GNN) showed different learning behavior. After exploring the solution space for only a few iterations, the models quickly converged towards a stable solution. This early stage of training is characterized by massive jumps in performance caused by the non-stationary environment in which the different agents optimize their policies individually. This explorative phase comes to a sudden stop when an equilibrium is found where no agent benefits from adapting its policies without other agents simultaneously changing their policies. Where multiple equilibriums exist, agents need to agree on one, which can result in the globally optimal solution or a local optimum.

\paragraph{}
Finally, IMARL’s learning curve directly reflects the model’s design. As all agents were initialized with a performant policy, the model achieved good results from the start. The first phase of training shows several short drops in performance, which mark the points where a new agent is starting its first training. Due to the reduced problem and solution spaces in IMARL, all agents quickly find a decent policy that brings overall model performance back to the previous level. Once all agents have been trained at least once, their policies are finetuned until an equilibrium is reached and training is stopped.

\subsection{IMARL Dependencies and Limitations}
\paragraph{}
The described learning patterns provide a visual explanation for IMARL’s superior performance compared to the other four models. While single-agent models need to interact in large problem and solution spaces that grow with increasing scenario complexity, IMARL benefits from the reduced dimensionality of the multi-agent approach. Where MARL and MARL+GNN suffer from non-stationarity, which can trap them in non-optimal equilibriums, IMARL finds superior solutions by removing the non-stationarity caused by the cyclic dependency of different agents and by initializing agents with a performant base policy.

\begin{figure}
  \centering
  \includegraphics[width=\textwidth,height=\textheight,keepaspectratio]{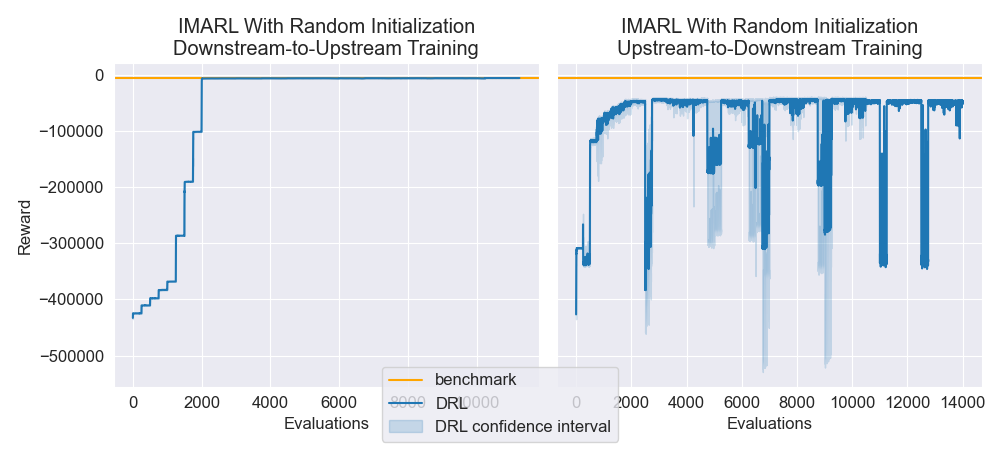}
  \caption{Example Learning Curves for IMARL With Random Initialization\\\textit{Comparison of learning curves for IMARL with random base policy and agents initially trained in downstream-to-upstream (left) and upstream-to-downstream (right) order.}}
  \label{fig:fig9}
\end{figure}

\paragraph{}
Using a base policy in IMARL to simulate a stable environment raises the question of how much IMARL's results depend on the quality of that policy and how applicable IMARL is to supply chain scenarios where no effective base policy is known. To investigate this question, we ran further experiments for IMARL without initializing agents based on a known policy. Instead, untrained agents take deterministic actions based on the initial weights of their respective actor-critic networks. The results of these experiments are shown in Figure \ref{fig:fig9}.

\paragraph{}
Even without random initialization, training time can be a challenge in IMARL. While the model successfully addresses the dimensionality and non-stationarity challenges of traditional SARL and MARL models, the sequential training approach requires significantly more episodes to reach convergence. Additional research can help find ways to reduce IMARL’s training time, for instance, by optimizing the training sequence and the duration of training individual agents.

\subsection{Review of Learned Policies}
\begin{figure}
  \centering
  \includegraphics[width=\textwidth,height=\textheight,keepaspectratio]{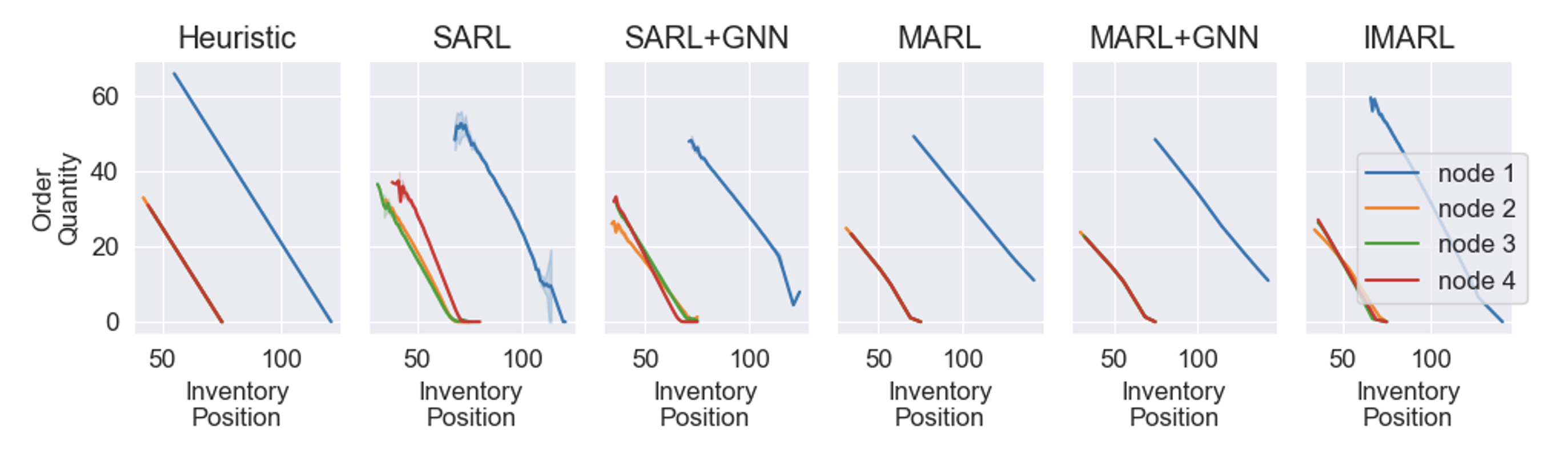}
  \caption{Learned Policies by Model\\\textit{Policies learned for each of the four stock points in complexity scenario A1 by the five DRL models and comparison with the benchmark heuristic.}}
  \label{fig:fig10}
\end{figure}

\paragraph{}
After having discussed experiment results with a focus on model performance and potential dependencies and limitations, we now review the policies that the models learned. As discussed in previous chapters, a policy in RL is defined by mapping states to actions. We visualized the policy learned by the model by simulating the supply chain over 100,000 time periods and recording the state and the corresponding actions selected by the model. Figure \ref{fig:fig10} visualizes the resulting policies for complexity scenario A1. The learned policies for all other scenarios are shown in Appendix D.

\paragraph{}
All models converged to policies that partially resemble the benchmark heuristic. As IP levels increase, replenishment order quantities strictly decrease until they reach zero. However, all policies differ slightly in terms of the rate by which order quantities decrease and the IP level for which no more units will be ordered. Additionally, for very low or very high IP values, most models exhibit some variance in their policies. Since these extreme state values occur less frequently in simulations, the models gain limited experience with such cases, leading to less clearly defined policies.

\newpage

\section{Conclusion}
\paragraph{}
In this paper, we examined the applicability of DRL for solving MEIO problems of real-world complexity. Although existing research has demonstrated that DRL can yield better results than commonly used heuristics, it is based on assumptions and simplifications that limit its applicability to supply chains in the real world. We have replicated the DRL model used by \cite{geevers2024} %Geevers et al. (2024) 
with some modifications and developed four enhanced models that utilize GNNs, MARL, and the novel IMARL approach. With the help of a complexity grid framework developed by us and using real-world demand and shipment lead time data, we simulated different supply chain scenarios to test the five DRL models. To assess the quality of these models, we compared their performance to that of a commonly used DA heuristic (\cite{rong2017}), %(Rong et al., 2017), 
which we generalized to the supply chain scenarios of our complexity grid. Building on the analysis presented in the previous chapter, this section synthesizes the key findings and their broader implications.

\subsection{Findings and Summary}
\paragraph{}
As expected, based on the findings from \cite{geevers2024}, %Geevers et al. (2024), 
our base model achieved good results in solving the MEIO problem for supply chain scenarios with lower complexity. When increasing the scenario complexity in our experiments by adding a third supply chain echelon, the base model can no longer beat the benchmark heuristic. Due to the curse of dimensionality, which is a known challenge for SARL, the larger problem and solution spaces keep the base model from finding superior solutions. Moreover, we observed a high variability of results, making it a less reliable method even for supply chain scenarios of limited complexity.

\paragraph{}
We addressed the dimensionality challenges of SARL by introducing a MARL-based model in which an RL agent represents every stock point in the supply chain network. As every agent needs to take an action only concerning its respective stock point, the solution space is reduced to a single dimension. Consequently, the MARL model can handle more complex scenarios than the SARL model and beats the benchmark for most networks with two echelons. As the number of nodes and echelons in the supply chain increases, MARL does not perform as well. This is due to the problem’s non-stationarity and the existence of multiple equilibriums that agents can agree on. Once an equilibrium is found, the solution tends to be stable, which helps MARL to achieve more consistent and reliable results than SARL.

\paragraph{}
To further address problem dimensionality, and non-stationarity, we enhanced both SARL and MARL models with GNNs. While we expected that GNNs would leverage the inherent graph structure of the supply chain network to improve collaboration between stock points and agents, our experiments showed that the GNN-based models performed at a similar level as their SARL and MARL counterparts in most scenarios. We hypothesize that the limited number of node features in our supply chain design makes it less suitable for GNNs and that they could add more value in other settings.

\paragraph{}
Based on these findings, we proposed IMARL, which addresses the non-stationarity issue inherent to MARL by training agents sequentially rather than simultaneously. IMARL consistently beat the benchmark heuristic in all supply chain scenarios along the complexity grid framework, making it an effective and reliable model. Although IMARL tends to reach reasonable solutions at an early stage of training, the sequential training approach paired with continuous retraining for best performance can take a long time to complete.

\subsection{Conclusion}
\paragraph{}
To our knowledge, our work is the first to test DRL for MEIO in supply chain scenarios of higher complexity. Circling back to our initial research question of whether DRL can be applied to MEIO problems of real-world complexity, we can conclude the following: Traditional DRL approaches like SARL and MARL have limited applicability to complex MEIO problems due to the dimensionality and non-stationarity challenges that are inherent to them. The iterative training approach proposed in IMARL addresses these challenges and makes it a viable option for solving complex MEIO problems. As the design of our simulated supply chain is not free from assumptions and simplifications, further research is required to test IMARL’s performance in even more realistic settings.

\subsection{Implications}
\paragraph{}
The proposed IMARL approach mitigated non-stationarity challenges, making it relevant to other fields of data science beyond inventory optimization. While a good base policy can help accelerate the training process in IMARL, it is not required by the model to perform well. This makes it a potential solution for problems where no such policy is available.

\paragraph{}
In most scenarios, IMARL achieved 6\%-14\% cost savings over the benchmark. This makes it relevant in supply chain management, where it can help businesses reduce costs while better serving their customers. It can also foster environmental sustainability by reducing excess inventory waste and avoiding emissions from overproduction and unnecessary shipments.

\paragraph{}
Lastly, the generalized DA heuristic served as a valuable indicator of the quality of the tested DRL models. Its efficiency, combined with its versatility in applying it to various network structures, demand patterns, and shipment lead time distributions, makes it highly relevant for organizations needing fast re-assessments of their inventory management policies.

\subsection{Limitations and Future Research}
\paragraph{}
Although this work might be the first to test DRL for MEIO in complex supply chain scenarios, it still relies on several assumptions and simplifications. Given the encouraging results of our experiments, additional research should be conducted to test the outlined methods in even more complex scenarios and to advance them further. In this section, we suggest several topics for future research that will contribute to making DRL a viable alternative to heuristics in inventory optimization of real-life supply chains.

\subsubsection{Additional Levels of Complexity}
\paragraph{}
As indicated in Chapter 3, we limited the scope of our research to 13 out of 30 supply chain scenarios as defined in the complexity grid framework. Running experiments for the remaining complex scenarios would mark a first step in further exploring the applicability of DRL to MEIO. In addition to testing new scenarios, we suggest modifying the logic by which stock points place replenishment orders in case of multiple upstream suppliers. While in our experiments, stock points randomly select a single upstream supplier from which they order the entire replenishment, a more realistic approach would be to allow order splitting and to choose internal suppliers based on stock availability.

\paragraph{}
As these changes to the environment would violate the conditions for which the DA heuristic was defined (\cite{rong2017}), %(Rong et al., 2017), 
a new benchmark heuristic must be developed. 

\subsubsection{Non-Stationary Changes to the Environment}
\paragraph{}
While our experiments used real-world data to simulate demand and shipment lead times, we removed non-stationarity from distributions by converting the original time series data into probability mass functions from which unlimited new time series can be generated. This approach requires re-training if significant changes to these distributions occur. For example, in the case of seasonal demand data, different models might be necessary to consider different demand patterns.

\paragraph{}
Apart from adding non-stationarity to the network parameters, the network structure could also be subject to changes. Simulating supply chain disruptions that make specific stock points or network connections unavailable for several periods would help test model resilience to external shocks common in global supply chains.

\paragraph{}
While the tested GNN-based models added limited value in the simulated supply chain scenarios, their usefulness for dealing with added non-stationarity should be tested.

\subsubsection{Transfer Learning in IMARL}
\paragraph{}
The previously discussed simulation of additional layers of non-stationarity also serves as an opportunity to test IMARL’s transfer learning capabilities. IMARL’s iterative training approach already relies on transfer learning, as agents are re-trained in a slightly changed environment where other agents have adapted their policies. It will be interesting to see whether IMARL can utilize its transfer learning capability to adjust to other environment changes, such as updated network parameters or modified network structures.

\subsubsection{Optimizing Efficiency and Effectiveness in IMARL}
\paragraph{}
As already indicated, IMARL can take a relatively long time to reach an equilibrium due to its sequential training approach combined with the continuous retraining of agents. By optimizing the duration of a single training iteration, additional research can help to improve the efficiency of IMARL’s training approach.

\paragraph{}
The order in which agents are trained also appears to influence both the duration of IMARL’s training and its results. Optimizing this training order—possibly with a greedy approach, in which agents with higher improvement potential are prioritized—could achieve further improvements.

\paragraph{}
By shortening IMARL’s training time and potentially enhancing its performance, it will become an even more valuable tool for addressing supply chain challenges, as well as broader data science problems.

\clearpage

\begin{appendices}
\section{List of Abbreviations}

\centering
\resizebox{0.5\textwidth}{!}{%
\begin{tabular}{ l l }
  \toprule
  Abbreviation & Meaning \\
  \midrule
  CPU & Central Processing Unit \\
  DA & Decomposition-Aggregation \\
  DRL & Deep Reinforcement Learning \\
  EOQ & Economic Order Quantity \\
  GAT & Graph Attention Network \\ 
  GNN & Graph Neural Network \\
  GPU & Graphics Processing Unit \\
  IPPO & Independent Proximal Policy Optimization (PPO) \\
  IMARL & Iterative Multi-Agent Reinforcement Learning \\
  IP & Inventory Position \\
  MARL & Multi-Agent Reinforcement Learning \\
  MAPPO & Multi-Agent Proximal Policy Optimization \\
  MDP & Markov Decision Process \\
  MEIO & Multi-Echelon Inventory Optimization \\
  PPO & Proximal Policy Optimization \\
  RL & Reinforcement Learning \\
  SARL & Single-Agent Reinforcement Learning \\
  SEIO & Single-Echelon Inventory Optimization \\
  \bottomrule
\end{tabular}%
}

\clearpage

\section{List of Hyperparameters}

\subsection{Global PPO Hyperparameters}
\centering
\resizebox{0.5\textwidth}{!}{%
\begin{tabular}{ l c }
  \toprule
  Hyperparameter & Value \\
  \midrule
  Network activation function & relu \\
  Learning rate for actor-critic network & 0.0001 \\
  Number of vectorized environments & 4 \\
  Number of environment steps per vectorized environment & 256 \\
  Number of PPO epochs per training data batch & 4 \\
  Number of minibatches per PPO epoch & 16 \\
  Discounting factor Gamma & 0.99 \\
  Lambda value for GAE computation & 0.95 \\
  Clipping value for PPO updates and value function & 0.2 \\
  Entropy regularization term for loss function & 0 \\
  Value function coefficient for the loss calculation & 0.5 \\
  Maximum norm of the gradients for a weight update & 0.5 \\
  \bottomrule
\end{tabular}%
}

\subsection{Model-Specific PPO Hyperparameters}
\centering
\resizebox{0.3\textwidth}{!}{%
\begin{tabular}{ l c }
  \toprule
  Hyperparameter & Value \\
  \midrule
  Actor-critic network size & \\
  \qquad SARL: & (256, 256) \\
  \qquad SARL+GNN: & (256, 256) \\
  \qquad MARL: & (64, 64) \\
  \qquad MARL+GNN: & (64, 64) \\
  \qquad IMARL: & (64, 64) \\
  GNN network size & (16, 16, 16) \\
  \bottomrule
\end{tabular}%
}

\subsection{Scenario- and Model-Specific PPO Hyperparameters}
\centering
\resizebox{0.6\textwidth}{!}{%
\begin{tabular}{ l c c c c c }
  \toprule
  Hyperparameter & SARL & SARL+GNN & MARL & MARL+GNN & IMARL \\
  \midrule
  Training episodes \\
  \qquad A1,A2: & 25,000 & 25,000 & 20,000 & 20,000 & 25,000/it \\
  \qquad A3,A4: & 75,000 & 75,000 & 20,000 & 20,000 & 25,000/it \\
  \qquad B1,B2: & 75,000 & 100,000 & 20,000 & 20,000 & 25,000/it \\
  \qquad B3,B4: & 150,000 & 200,000 & 20,000 & 20,000 & 25,000/it \\
  \qquad C1,C2: & 75,000 & 100,000 & 20,000 & 20,000 & 25,000/it \\
  \qquad C3,C4: & 150,000 & 200,000 & 20,000 & 20,000 & 25,000/it \\
  \qquad D1: & 300,000 & 300,000 & 20,000 & 20,000 & 25,000/it \\
  \bottomrule
\end{tabular}%
}

\section{Learning Curves by Model and Complexity Scenario}
\centering
\includegraphics[width=\textwidth,height=\textheight,keepaspectratio]{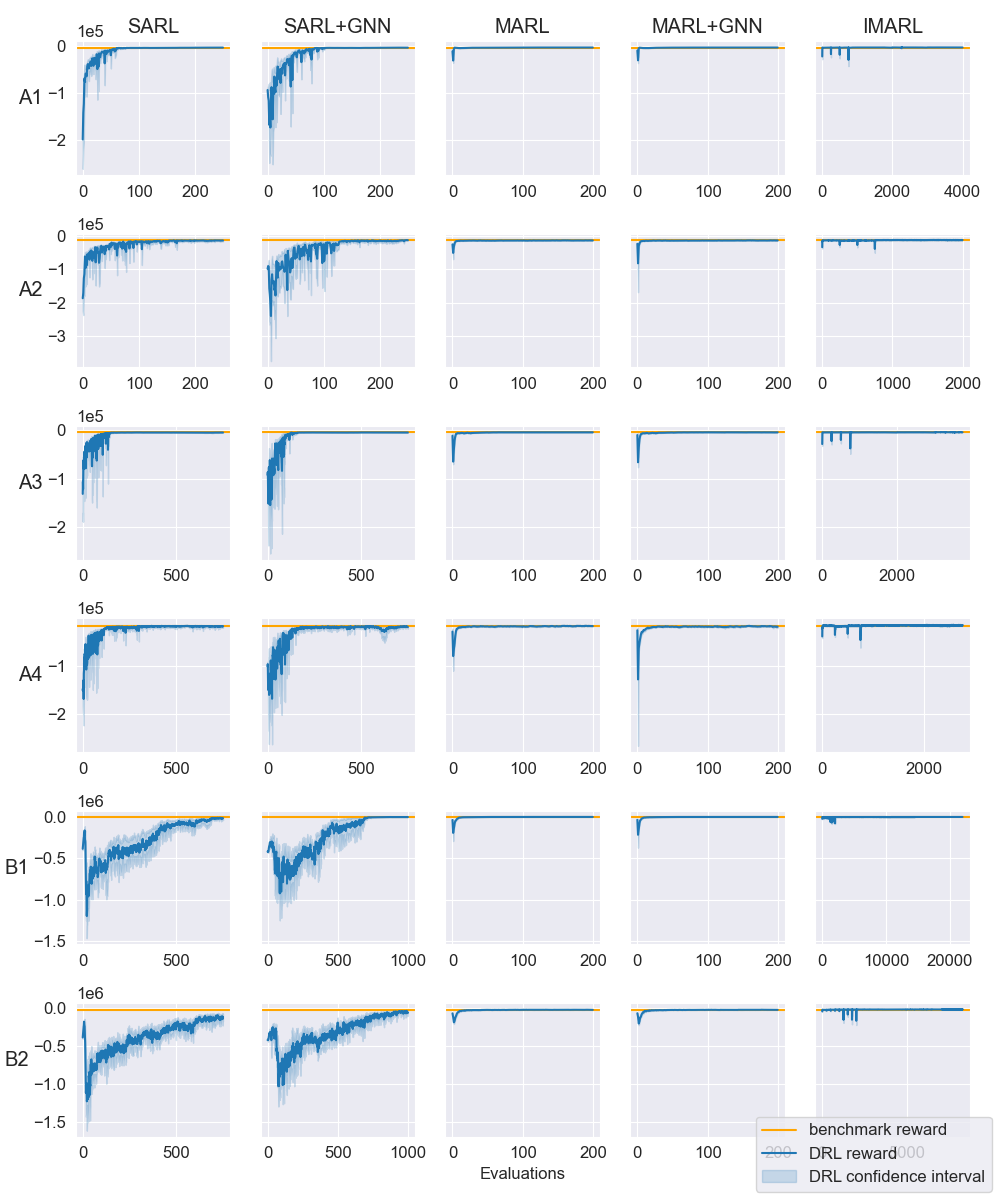}

\centering
\includegraphics[width=\textwidth,height=\textheight,keepaspectratio]{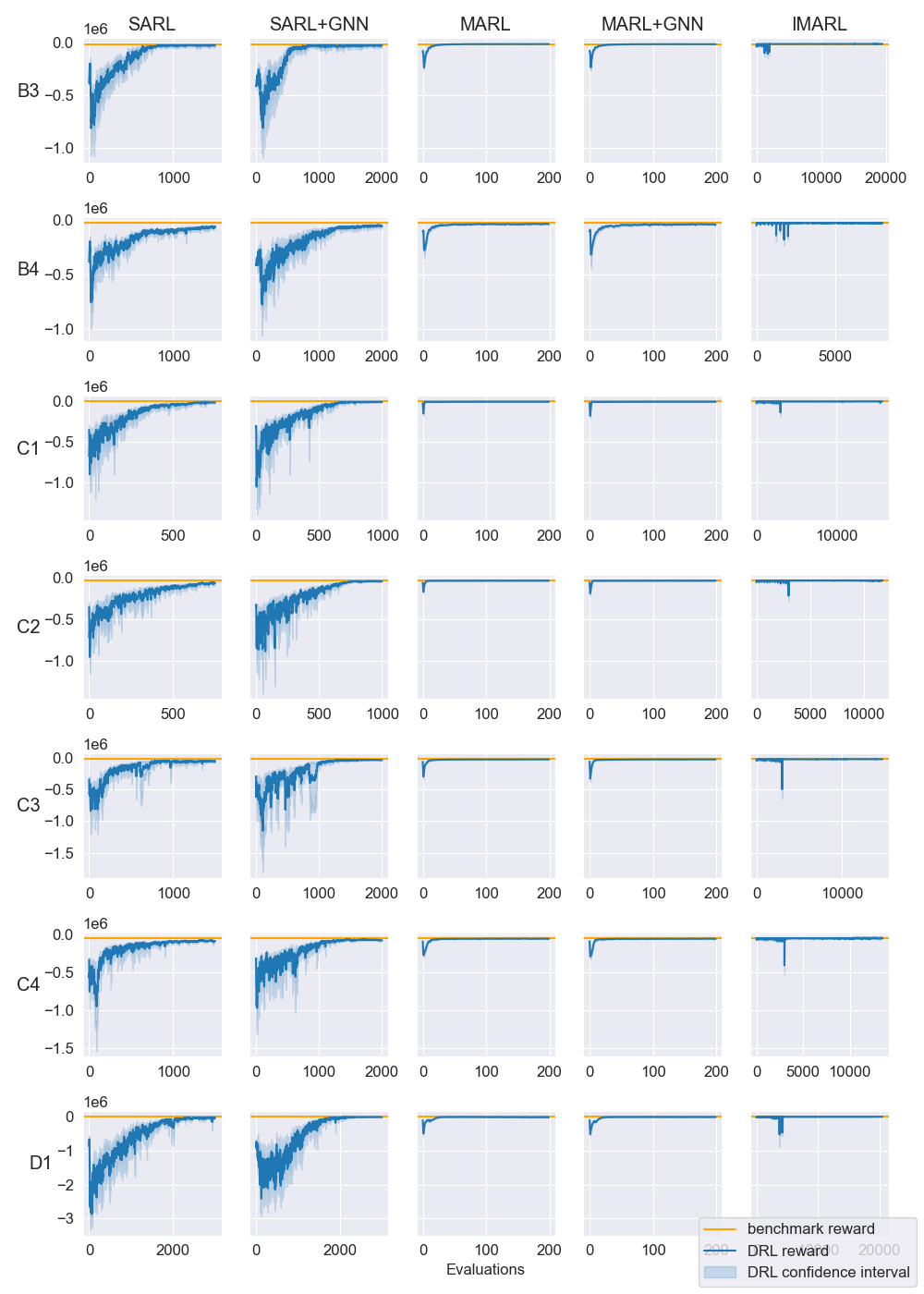}

\section{Learned Policies by Model and Complexity Scenario}
\centering
\includegraphics[width=\textwidth,height=\textheight,keepaspectratio]{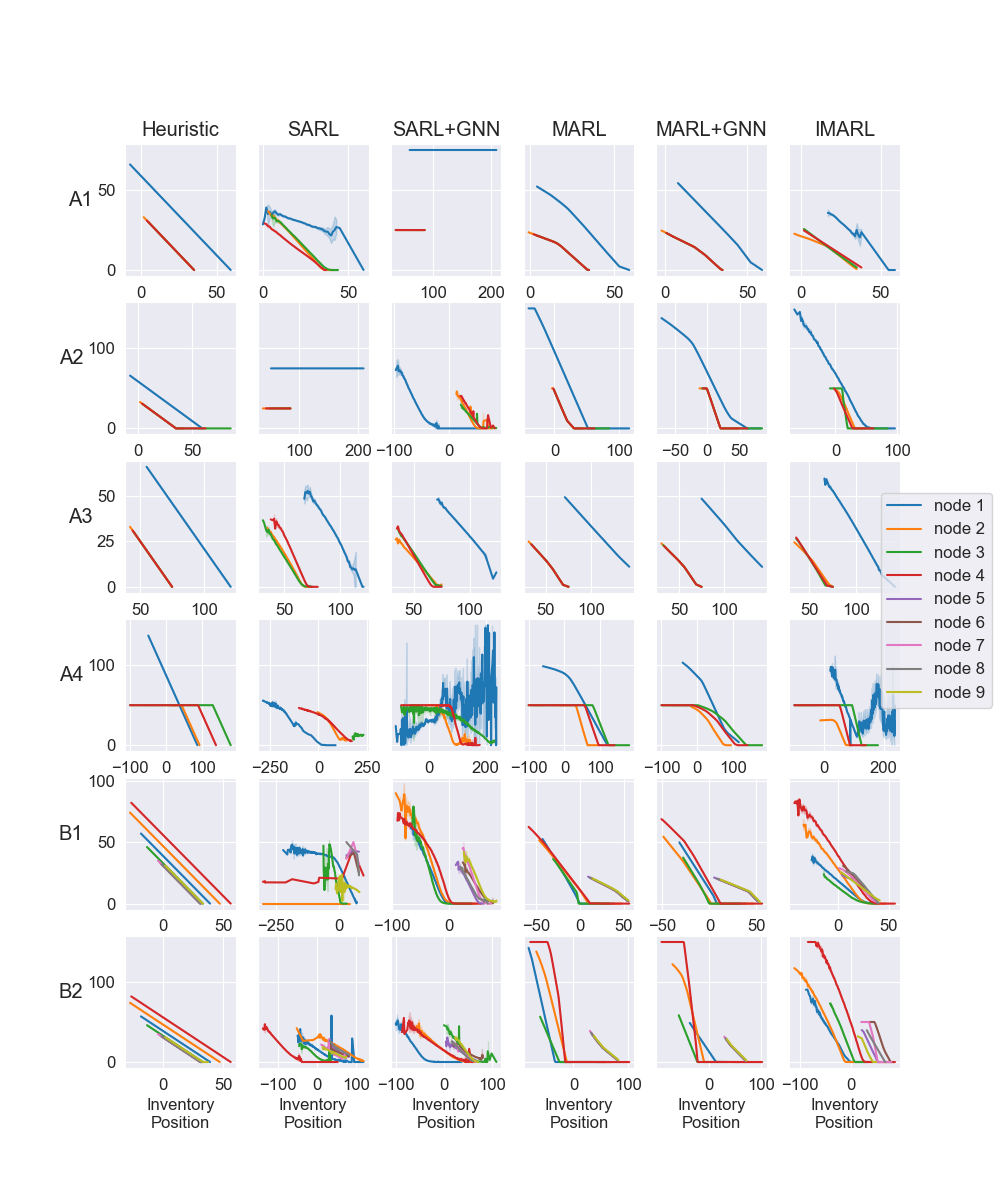}

\centering
\includegraphics[width=\textwidth,height=\textheight,keepaspectratio]{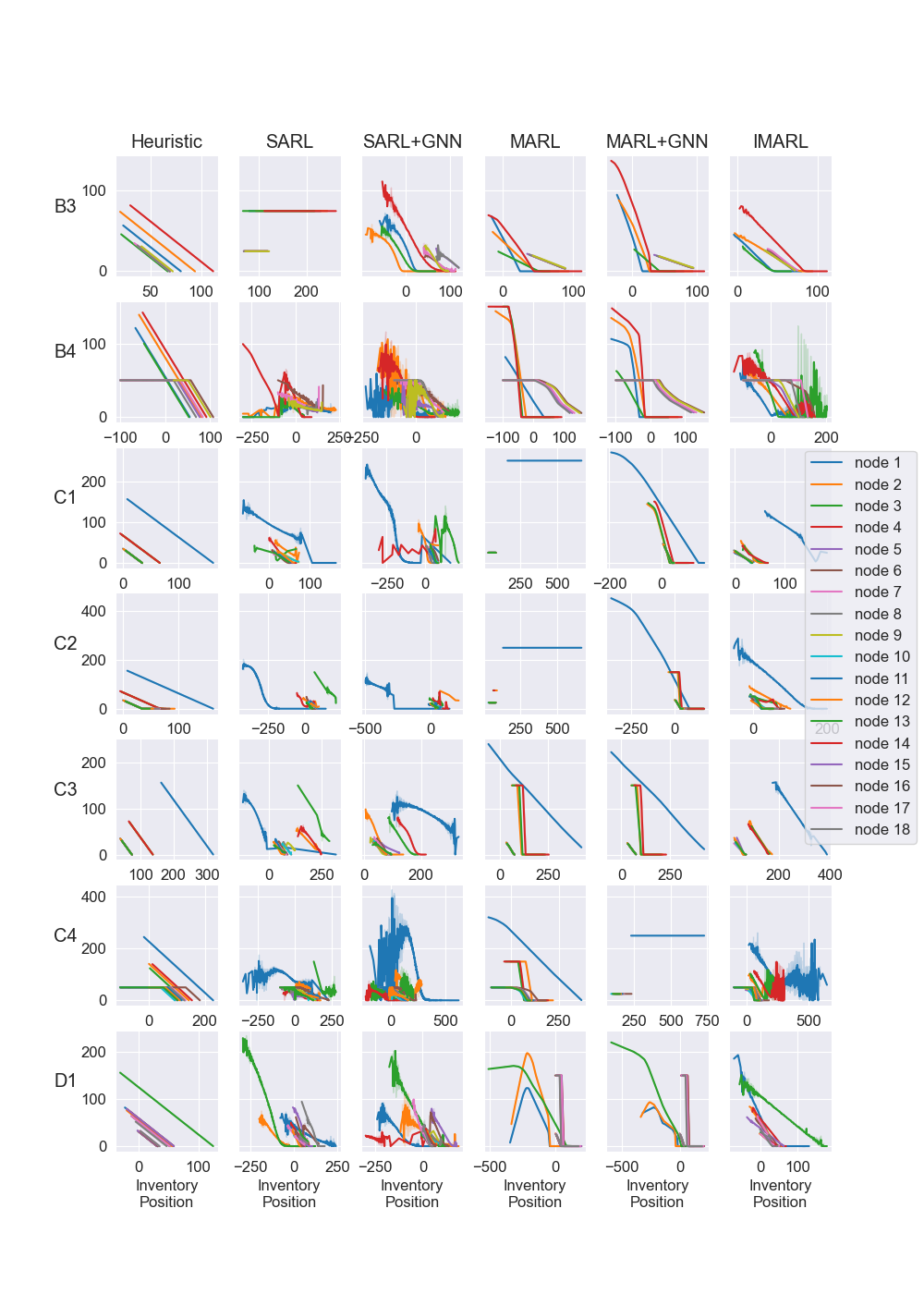}

\clearpage

\bibliographystyle{unsrtnat}
\bibliography{references}  %%% Uncomment this line and comment out the ``thebibliography'' section below to use the external .bib file (using bibtex) .


% George Kour and Raid Saabne.
% \newblock Real-time segmentation of on-line handwritten arabic script.
% \newblock In {\em Frontiers in Handwriting Recognition (ICFHR), 2014 14th International Conference on}, pages 417--422. IEEE, 2014.



% \end{thebibliography}

\end{appendices}
\end{document}